\documentclass[10pt,journal,compsoc]{IEEEtran}
\ifCLASSOPTIONcompsoc
  \usepackage[nocompress]{cite}
\else
  \usepackage{cite}
\fi
\ifCLASSINFOpdf
  \usepackage[pdftex]{graphicx}
\else
\fi
\usepackage{amsmath}
\usepackage{bbm}
\usepackage{algorithmic}

\usepackage{array}

\ifCLASSOPTIONcompsoc
  \usepackage[caption=false,font=footnotesize,labelfont=sf,textfont=sf]{subfig}
\else
  \usepackage[caption=false,font=footnotesize]{subfig}
\fi
\usepackage{fixltx2e}

\usepackage{stfloats}
\usepackage{url}

\usepackage{amsmath,amsfonts}
\usepackage{array}
\usepackage{booktabs}
\usepackage{multirow}
\usepackage{hyperref}
\hypersetup{colorlinks,linkcolor={red},citecolor={cyan},urlcolor={magenta}} 
\usepackage{breakcites}
\usepackage{threeparttable}
\usepackage{caption}
\usepackage[misc,geometry]{ifsym}
\usepackage[table]{xcolor} 
\usepackage[most]{tcolorbox}
\usepackage{ragged2e}
\usepackage{xspace}
\usepackage{makecell}

\definecolor{LightCyan}{rgb}{0.88,1,1}
\definecolor{Gray}{gray}{0.90}

\definecolor{forestgreen}{rgb}{0.133, 0.545, 0.133}
\definecolor{yellowyellow}{rgb}{0.133, 0.545, 0.133}
\definecolor{lightblue}{RGB}{68, 114, 196}
\definecolor{correct}{RGB}{173, 173, 173}
\definecolor{incorrect}{RGB}{234, 59, 46}

\makeatletter
\DeclareRobustCommand\onedot{\futurelet\@let@token\@onedot}
\def\@onedot{\ifx\@let@token.\else.\null\fi\xspace}

\def\eg{\emph{e.g}\onedot} 
\def\ie{\emph{i.e}\onedot} 
 
 \def\vs{\emph{vs}\onedot}

\def\aka{\emph{a.k.a}\onedot}
\makeatother

\newcommand{\mparagraph}[1]{\vspace{1mm}\noindent{\textbf{#1.}\hspace{1mm}}}
\renewcommand{\figurename}{Figure}
\usepackage{bm}
\usepackage[utf8]{inputenc}
\usepackage{amssymb}
\usepackage{tikz}
\usetikzlibrary{arrows.meta} 
\usetikzlibrary{calc}
\newcommand{\myhookarrow}{
  \tikz[baseline=-0.3ex,scale=0.15,line width=0.45pt]{%
    \draw[-{Stealth}] (0,1) -- (0,0) -- (2,0); 
  }
}

\usepackage{amssymb}

\begin{document}

\title{Scalable and Generalizable Correspondence Pruning via Geometry-Consistent Pre-training}

\author{Tangfei Liao, Xiaoqin Zhang\textsuperscript{$\ast$},~\IEEEmembership{Senior member,~IEEE}, Tao Wang, Hao Ye, Min Li, \\
Guobao Xiao,~\IEEEmembership{Senior member,~IEEE}, and Mang Ye\textsuperscript{$\ast$},~\IEEEmembership{Senior member,~IEEE}
\IEEEcompsocitemizethanks{
\IEEEcompsocthanksitem This work was supported by the National Natural Science Foundation of China under Grants T2541022, 62361166629, and U24A20242, the National Key Research and Development Program of China under Grant 2024YFC3306902, and the Major Project of Science and Technology Innovation of Hubei Province under Grant 2025BEA002. 
\IEEEcompsocthanksitem T. Liao and M. Ye are with the School of Computer Science, Wuhan University, Wuhan, China (E-mail: tangfeiliao@whu.edu.cn). 
\IEEEcompsocthanksitem X. Zhang and M. Li are with the College of Computer Science and Artificial Intelligence, Wenzhou University, Wenzhou, China (E-mail: limin.simu@gmail.com). 
\IEEEcompsocthanksitem T. Wang is with the State Key Lab for Novel Software Technology, Nanjing University, Nanjing, China (E-mail: taowangzj@gmail.com). 
\IEEEcompsocthanksitem H. Ye is with the Xiaomi Corporation, Xiaomi Automobile Co., Ltd., Beijing, China (E-mail: yehao5@xiaomi.com). 
\IEEEcompsocthanksitem G. Xiao is with the School of Computer Science and Technology, Tongji University, Shanghai, China (E-mail: x-gb@163.com). 
\IEEEcompsocthanksitem \textsuperscript{$\ast$}Corresponding authors are Xiaoqin Zhang and Mang Ye: zhangxiaoqinnan@gmail.com, yemang@whu.edu.cn. 
}}

\IEEEtitleabstractindextext{
\begin{abstract}
\RaggedRight 
\justifying
Two-view correspondence pruning aims to identify reliable correspondences for camera pose estimation, serving as a fundamental step in many 3D vision tasks. 
Existing methods rely on geometric consistency to seek true correspondences (inliers) from numerous false correspondences (outliers). 
In this learning paradigm, outliers severely affect the representation learning of inliers, resulting in models that are neither robust nor generalizable. 
To address this issue, we propose a geometry-consistent pre-training paradigm that sculpts scalable and generalizable representations free from outlier interference. 
The paradigm features two appealing properties. 
\textbf{1) Implementation of geometry-consistent pre-training.} 
We introduce masked inlier reconstruction as a pretext task and develop a simple yet effective pre-training framework based on a masked autoencoder. 
Specifically, due to the irregular and unordered nature of correspondences, which lack explicit positional information, we adopt a dual-branch structure that separately reconstructs the keypoints of two images. 
This enables indirect reconstruction of 4D correspondences, where keypoints from the paired image provide positional prompts. 
\textbf{2) Unified correspondence encoder.} 
We propose a simple dual-stream encoder with built-in consensus interaction, providing a unified, extensible architecture that enhances representation learning. 
Extensive experiments demonstrate that our method, GeneralPruner, consistently outperforms state-of-the-art approaches in terms of robustness and generalization across various downstream tasks.
Specifically, our method achieves $10.76$\%, $11.84$\%, and $8.65$\% performance gains in camera pose estimation, visual localization, and 3D registration, respectively.
To the best of our knowledge, we are the first work to introduce a pre-training framework tailored for correspondence pruning, offering a more universal and scalable solution. 
Code is publicly available at \url{https://github.com/sugar-fly/GeneralPruner}. 
\end{abstract}

}

\maketitle

\IEEEdisplaynontitleabstractindextext
\IEEEpeerreviewmaketitle
\IEEEraisesectionheading{\section{Introduction}\label{sec:introduction}}

\IEEEPARstart{F}{inding} reliable geometric correspondences for two-view camera pose estimation is the cornerstone of many 3D computer vision applications, including 3D reconstruction~\cite{agarwal2011building, schonberger2016structure, kerbl3Dgaussians}, SLAM systems~\cite{campos2021orb, mur2017orb, mur2015orb}, and visual localization~\cite{sarlin2019coarse, sarlin2021back}. 
Geometric matching can be broadly grouped into sparse, semi-dense, and dense paradigms. 
Sparse pipelines detect keypoints and match descriptors~\cite{Lowe2004, DeTone2018, Sarlin2020}, 
while recent semi-dense and dense approaches adopt detector-free or end-to-end formulations~\cite{sun2021loftr, truong2021learning, edstedt2023dkm} to predict correspondences directly. 
Despite impressive progress, all these paradigms can still produce many false correspondences (\aka outliers) under challenging conditions such as large viewpoint changes, low texture, repetitive patterns, or cross-modal settings~\cite{ma2021image}. 
Hence, considerable recent research has paid attention to correspondence pruning, aiming to accurately differentiate geometry-consistent correspondences (\aka, inliers) from false ones and improve camera pose estimation. 
This paper also focuses on such techniques to improve the performance of downstream tasks. 

Traditional approaches have predominantly followed the hypothesize-and-verify paradigm, with algorithms such as RANSAC~\cite{Fischler1981} and its variants~\cite{Chum2005a, Raguram2012, Barath2020} achieving widespread success. 
These methods typically operate under predefined parametric model constraints, where a subset of putative correspondences is first sampled, 
and the model is subsequently refined through iterative verification to maximize the number of inlier supporters. 
However, such handcrafted methods often exhibit limited robustness in the presence of dominant outliers~\cite{ma2021image}, leading to poor generalization. 
Thus, the perennial theme of two-view correspondence pruning lies in obtaining more generalized models. 

To achieve this goal, recent advances resort to deep networks~\cite{Yi2018, Zhang2019, liu2021learnable, zhong2021t, convmatch2023, li2024u, liu2023ncm, dai2024mgnet, miao2024bclnet} to learn a robust model generalizable to unseen scenes from training scenes. 
The most popular models commonly employ iterative networks, where each iteration inherits weights from the previous one, progressively refining correspondence representations. 
These methods primarily design diverse graph-based encoders to capture inlier consistency, as exemplified by OANet~\cite{Zhang2019}, CLNet~\cite{Zhao2021}, LMCNet~\cite{liu2021learnable}, and VSFormer~\cite{liao2023vsformer}. 
However, the learning process of correspondence pruning is significantly affected by outliers, leading to weak inlier representations and severely limiting model generalization. 
It also remains an open question how to learn a model that achieves stable generalization across scenes and domains. 

\begin{figure*}[t]
\begin{center}
    \includegraphics[width=1\linewidth]{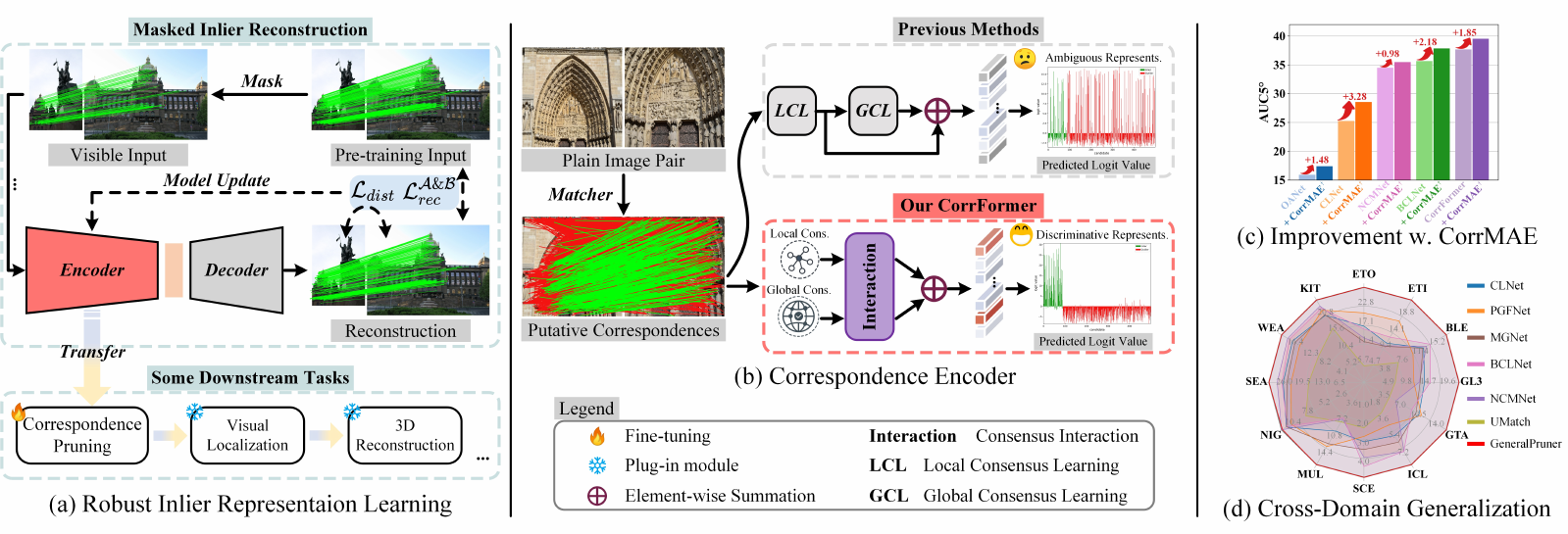}
    \caption{Overview of our GeneralPruner. (a) The proposed masked inlier reconstruction. 
    The goal of this task is to effectively reconstruct true correspondences by leveraging inlier geometric consistency, then transfer the robust and generalizable representation for downstream tasks. 
    (b) Our encoder features a simple dual-stream architecture with built-in consensus interaction, allowing it to capture multi-scale dependencies. 
    (c) Our encoder-agnostic pre-training method, can be seamlessly integrated to improve several SoTA pruners~\cite{Zhang2019, Zhao2021, liu2023ncm, miao2024bclnet}. 
    (d) SoTA pruners~\cite{Zhao2021, liu2023pgfnet, li2024u, liu2023ncm, dai2024mgnet, miao2024bclnet} struggle with the zero-shot out-of-domain benchmark~\cite{xuelun2024gim}, while our method greatly improves generalization.}\label{fig:overview}
\end{center}
\vspace{-0.5cm}
\end{figure*}

In contrast, we shift our attention to the trending pre-training methodologies to endow correspondence pruning with generalizable representations. 
Recent works~\cite{pang2022masked, tong2022videomae, chen2023traj} have demonstrated that simple masked modeling~\cite{He2022mae, devlin2018bert} can substantially improve performance across diverse downstream tasks. 
Although masked modeling is widely used in NLP and CV, it has not been explored in correspondence pruning. 
In this paper, we are the first to apply masked modeling to this field. 
Our goal is to learn geometry-consistent representations that generalize well to downstream tasks. 
Specifically, we introduce a pretext task called \textit{masked inlier reconstruction} for correspondence pruning, as illustrated in Figure~\ref{fig:overview}(a). 
That is motivated by two main considerations: 
\textbf{1) Robustness.} Since masked inlier reconstruction is conducted without the interference of outliers, it allows the model to focus solely on learning geometry-consistent inlier representations. 
This leads to more robust geometry-consistent representations with improved generalization. 
\textbf{2) Scalability.} Normally, training correspondence pruning models needs accurate geometric labels from structure-from-motion/multi-view stereo~\cite{li2018megadepth}, which is complex and costly. 
Our pretext task does not rely on these labels. 
We can generate synthetic two-view correspondences using simple geometric matching methods~\cite{edstedt2023dkm, miao2024bclnet} and use abundant Internet videos for scalable and cost-effective pre-training. 

Based on the above analysis, a naive implementation of this pretext task entails employing a popular framework, Masked Autoencoder (MAE)~\cite{He2022mae}, to reconstruct masked inliers directly, which is similar to Point-MAE~\cite{pang2022masked}. 
Unfortunately, unlike point clouds, correspondences lack a well-defined geometric center or other explicit location information for position encoding. 
That is, this solution ignores the unordered and irregular characteristics of correspondence, leading to ineffective reconstruction of masked inliers. 
To this end, we extend MAE and propose a novel pre-training framework, \textit{CorrMAE}. 
The core idea of CorrMAE involves a dual-branch structure that separately recovers masked keypoints in the source and target images. 
It also leverages keypoints from the other image as positional prompts, indirectly enabling the reconstruction of masked inliers. 

With the aforementioned designs, various correspondence encoders~\cite{Zhang2019, Zhao2021, liu2023ncm, miao2024bclnet}—serving as bridges between pre-training and correspondence pruning—can be seamlessly integrated into CorrMAE, 
thus enhancing their downstream performance through transfer learning, as illustrated in Figure~\ref{fig:overview}(c). 
However, the encoder itself becomes the performance bottleneck, since existing designs lack explicit mechanisms for consensus-level interaction and therefore fail to capture multi-scale consensus dependencies. 
To address this limitation, we propose \textit{CorrFormer}, a more general and powerful encoder tailored for correspondence learning, as illustrated in Figure~\ref{fig:overview}(b). 
CorrFormer adopts a dual-stream architecture that separately models local and global consensus, while enabling implicit interaction through weight-shared linear transformers. 
This design not only facilitates deeper geometric reasoning across correspondences but also provides a unified and scalable architecture compatible with the pre-training. 
When coupled with CorrMAE, the resulting model, termed \textbf{GeneralPruner}, further demonstrates significant improvements in zero-shot cross-domain generalization, as shown in Figure~\ref{fig:overview}(d).

To summarize, our key contributions are as follows:
\begin{itemize}
	\item We propose a geometry-consistent pre-training paradigm for correspondence pruning, 
    which eliminates outlier interference during representation learning and enables scalable pre-training without geometric supervision. 
    To our knowledge, this is the first work to improve the generalization of pruners from a pre-training perspective. 
    \item We implement this paradigm with CorrMAE, a simple yet effective pre-training framework based on the MAE. 
    It adopts a dual-branch design that reconstructs keypoints of paired images, enabling effective correspondence reconstruction. 
    The framework is encoder-agnostic and can be seamlessly integrated with various encoders, further boosting their performance through transfer learning. 
	\item We propose CorrFormer, a unified and extensible dual-stream encoder with built-in consensus interaction for correspondence learning. 
    This design jointly models local and global consensus and facilitates implicit feature interaction, providing a more general and powerful representation architecture. 
    \item Extensive experiments demonstrate that GeneralPruner achieves new state-of-the-art performance across multiple downstream tasks, 
    with improvements of $10.76$\% in camera pose estimation, $11.84$\% in visual localization, and $8.65$\% in 3D registration. 
\end{itemize}

\section{Related Work}
Image matching broadly includes sparse, semi-dense, and dense paradigms. 
Sparse methods detect keypoints (\eg, SIFT~\cite{Lowe2004}, SuperPoint~\cite{DeTone2018}) and use learnable matchers or mutual nearest-neighbor, to predict correspondences~\cite{Sarlin2020, chen2021learning, xue2023imp}. 
Semi-dense approaches refine correspondences in a coarse-to-fine manner~\cite{sun2021loftr, jiang2021cotr, chen2022aspanformer}, 
while dense methods estimate a dense warp with confidence for sampling correspondences~\cite{truong2021learning, edstedt2023dkm, zhu2023pmatch}. 
Despite strong accuracy, end-to-end semi/dense models can become computationally expensive at high resolutions~\cite{lindenberger2023lightglue}, whereas sparse pipelines remain attractive for time-critical settings such as SLAM. 
Nevertheless, all paradigms may produce many outliers under wide baselines, illumination changes, repetitive structures, or cross-modal sift, which has spurred increasing interest in correspondence pruning. 
We next review advances in correspondence pruning and masked modeling. 

\subsection{Correspondence Pruning}

\mparagraph{Traditional Methods}
RANSAC~\cite{Fischler1981} and its variants~\cite{Chum2005a, Raguram2012, brachmann2017dsac, brachmann2019neural, Barath2019, Barath2020} iteratively apply the hypothesize-and-verify strategy to find an optimal geometric model, \eg, essential matrix represented relative camera pose. 
To be specific, these methods first sample a subset of putative correspondences to estimate a parametric model via the five-point algorithm~\cite{nister2004efficient}, which serves as a model hypothesis. 
Then, they iteratively refine the model until it maximizes the number of supporters. 
Among them, PROSAC~\cite{Chum2005a} leverages the prior information (\eg, feature matching scores) of correspondences to optimize the sampling process, prioritizing high-confidence samples for hypothesis generation. 
This accelerates the hypothesize-and-verify process. 
USAC~\cite{Raguram2012} introduces a modular framework that integrates various optimization strategies. 
MAGSAC~\cite{Barath2019} reduces the sensitivity of fixed thresholds to noise. 
It achieves this by marginalizing the uncertainty of the inlier threshold. 
Building on that, MAGSAC++~\cite{Barath2020} further enhances computational efficiency and model selection strategies. 
However, as the proportion of outliers in putative correspondences increases, these methods exhibit limited robustness, poor generalization, and low computational efficiency, making them challenging to apply in complex scenarios with a large number of outliers~\cite{ma2021image}. 

\mparagraph{Correspondence Learning}
In recent years, correspondence learning has emerged as a powerful tool for two-view correspondence pruning. 
A range of methods have shown promising performance in the presence of numerous outliers. 
PointCN~\cite{Yi2018}, as a pioneering work, formulates correspondence pruning as both a binary classification problem (\ie, outlier rejection) and an essential matrix regression problem (\ie, camera pose estimation). 
It identifies inliers using geometric consistency constraints and estimates the essential matrix encoding the camera pose via a weighted eight-point algorithm~\cite{hartley1997defense}. 
Moreover, it employs a permutation-equivariant network with context normalization~\cite{ulyanov2016instance} as the correspondence encoder. 

Subsequent works design structures or manners to enhance correspondence encoders. 
OANet~\cite{Zhang2019} builds an implicit graph with soft assignment to capture local consensus, while CLNet~\cite{Zhao2021} extends to local-to-global consensus using DGCNN/GCN~\cite{wang2019dynamic, kipf2016semi}. 
NCMNet~\cite{liu2023ncm} enriches local consensus via multiple subspaces and BCLNet~\cite{miao2024bclnet} further improves it with bilateral attention. 
Alternatively, ConvMatch~\cite{convmatch2023} maps correspondences to a motion vector field and mine motion consistency using 2D convolutions. 
Despite progress, learning discriminative inlier representations under massive outliers remains difficult, often harming generalization to cross-domains. 
We address this with geometry-consistent pre-training, where masked inlier reconstruction avoids outlier interference to enable low-cost data scaling, and a simple extensible encoder aligned with this framework further improves generalization. 

\subsection{Masked Modeling}
BERT~\cite{devlin2018bert} is a seminal language representation model that leverages masked token prediction to learn deep bidirectional contextual embeddings. 
MAE~\cite{He2022mae} extends this masked modeling paradigm to the visual field by reconstructing masked image patches, enabling scalable and effective self-supervised learning for visual representations. 
Building on this idea, subsequent studies have adopted the MAE framework for pre-training across a variety of tasks, including 3D object classification~\cite{pang2022masked}, video understanding~\cite{tong2022videomae}, trajectory prediction~\cite{chen2023traj}, and action recognition~\cite{yan2023skeletonmae}. 
However, MAEs fail to reconstruct correspondences due to their unordered and irregular characteristics. 
To this end, we extend MAE and leverage a dual-branch structure to reconstruct masked inliers, providing robust geometry-consistent representations for downstream tasks. 

\begin{figure*}[t]
\begin{center}
    \includegraphics[width=1\linewidth]{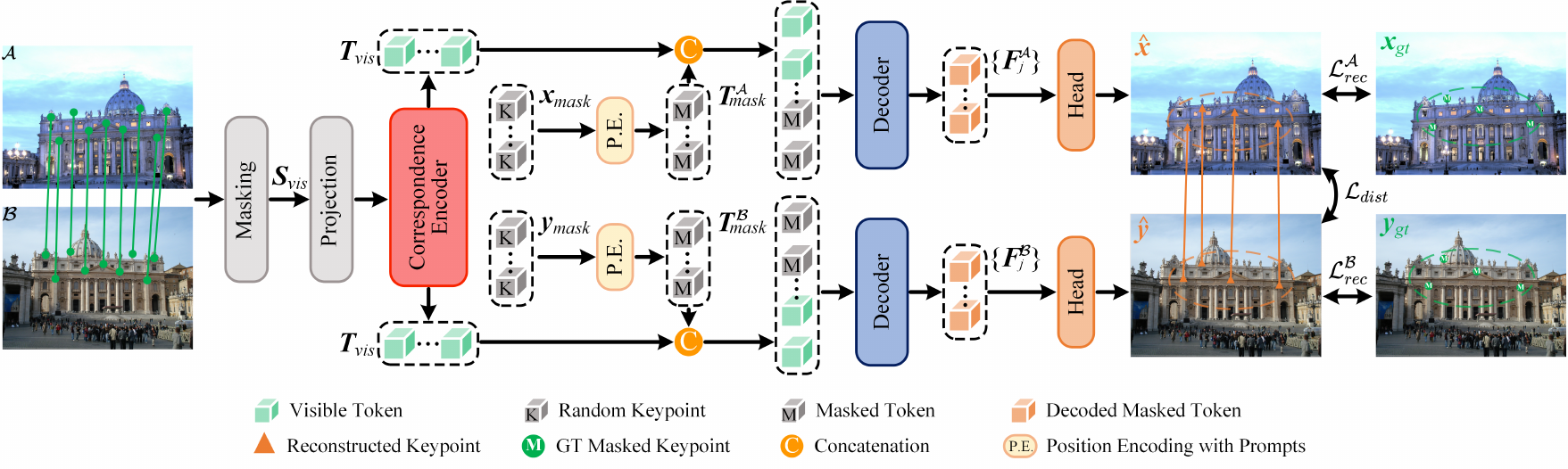}
    \caption{Details of our proposed CorrMAE. 
    Given a set of correspondences, either ground-truth or generated by a geometric matcher~\cite{xue2025matcha, xuelun2024gim}, CorrMAE indirectly recovering unordered correspondences in 4D space. 
    It aims to learn geometry-consistent representations through pre-training and transfer them to downstream tasks to enhance generalization.}\label{fig:corrmae}
\end{center}
\vspace{-0.4cm}
\end{figure*} 

\section{Method}
\subsection{Overview} 
Given a pair of images $(\mathcal{A}_\mathcal{F}, \mathcal{B}_\mathcal{F})$, an off-the-shelf matcher~\cite{Lowe2004, Sarlin2020} generates putative correspondences. 
The correspondence pruning model, or pruner, seeks to mine geometric consistency and learn discriminative representations from these correspondences. 
However, numerous outliers within the putative set severely limit the pruner’s robustness and generalization. 
A straightforward solution is to scale up training with fully supervised pruners, making them more general, similar to the pre-training of ResNet~\cite{he2016deep} on ImageNet~\cite{russakovsky2015imagenet}. 
Unfortunately, this requires a costly and complex structure-from-motion system~\cite{li2018megadepth} to generate annotations, which limits scalability. 
To overcome this, we propose a geometry-consistent pre-training paradigm for correspondence pruning. 
On one hand, we introduce masked inlier reconstruction as a pretext task, which learns robust and generalizable representations free from outlier interference. 
On the other hand, it does not rely on geometric supervision, with training data easily generated through simple geometric matchers~\cite{xue2025matcha, xuelun2024gim}, 
enabling the model to leverage large-scale Internet videos and diverse scene images (see Section~\ref{sec:general}). 
In practice, we extend MAE~\cite{He2022mae} and adopt a dual-branch structure to effectively implement masked inlier reconstruction. 
It can also seamlessly integrate with various correspondence encoders and further improve the performance of correspondence pruning. 
It is evident that the encoder emerges as the core component of our method, serving as the bridge between pre-training and fine-tuning. 
Thus, we revisit existing design choices and propose a dual-stream encoder with built-in consensus interaction, making it more extensible and further boosting its capacity. 

\begin{figure}[t]
\begin{center}
    \includegraphics[width=0.8\linewidth]{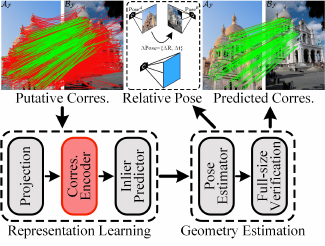}
    \vspace{-0.2cm}
    \caption{Pipeline of correspondence pruning. 
    The core part is the correspondence encoder, which mines inlier consistency to support binary classification and pose regression.}\label{fig:cp}
\end{center}
\vspace{-0.5cm}
\end{figure}

\mparagraph{Pre-training with CorrMAE}
We propose correspondence masked autoencoder (\textit{CorrMAE}), a pre-training framework that extends MAE~\cite{He2022mae} to perform masked inlier reconstruction, as illustrated in Figure~\ref{fig:corrmae}. 
Given $M$ two-view correspondences, either ground-truth or generated by a geometric matcher~\cite{xue2025matcha, xuelun2024gim}, 
\begin{gather} \label{eq1}
    \left\{ \bm{S}_{i}=(\bm{x}_i, \bm{y}_i)|i=1,...,M, \bm{x}_i \in \mathbb{R}^2, \bm{y}_i \in \mathbb{R}^2 \right\}, 
\end{gather}
those are selected from putative correspondences based on geometric labels~\cite{Hartley2003} or refined using a SoTA pruner. 
The reconstruction task begins by randomly sampling correspondences with a masking ratio. 
Subsequently, sampled visible correspondences are embedded in global and local consensus using the correspondence encoder $\mathcal{E}_\theta$. 
At last, guided by the consistency within visible correspondences, the decoder $\mathcal{D}_\theta$ conducts masked inlier reconstruction. 
More details will be described in Section~\ref{sec:corrmae} and Section~\ref{sec:loss}. 

\mparagraph{Fine-tuning for Correspondence Pruning}
Given $N$ putative correspondences, 
\begin{gather} \label{eq2}
    \left\{ \bm{C}_{i}=(\bm{u}_i, \bm{v}_i)|i=1,...,N, \bm{u}_i \in \mathbb{R}^2, \bm{v}_i \in \mathbb{R}^2 \right\}. 
\end{gather} 
Correspondence pruning is typically formulated as both a binary classification problem (\ie, inlier vs. outlier) and an essential matrix regression problem~\cite{Yi2018}. 
However, identifying geometry-consistent true correspondences from an outlier-dominated putative set remains challenging. 
To alleviate this challenge, we propose a dual-stream encoder (see Section~\ref{sec:encoder}) pretrained via CorrMAE (see Section~\ref{sec:corrmae}). 
As illustrated in Figure~\ref{fig:cp}, each correspondence passes through a projection layer and an encoder, after which the inlier predictor (\ie, multiple MLPs) outputs a logit set $\bm{\hat{w}} = {\left\{ \hat{w_i} \right\}}$. 
A pose estimator, such as the weighted eight-point algorithm~\cite{hartley1997defense, Yi2018}, is then employed to estimate the essential matrix $\bm{\widehat{E}}$, which encodes the relative camera pose. 
The process can be expressed as, 
\begin{align} \label{eq3}
    \bm{\hat{w}} &= \mathrm{MLP} (\mathcal{E}_{\theta}(\mathrm{Proj} (\bm{C}))), \\
    \bm{\widehat{E}} &= g(\bm{C}, \mathrm{Softmax}(\bm{\hat{w}})). 
\end{align}
Meanwhile, we adopt iterative pruning networks for fine-tuning, which is the same as CLNet~\cite{Zhao2021} and NCMNet~\cite{liu2023ncm}. 
Further details on the supervision of iterative networks are provided in Section~\ref{sec:loss}. 
In addition, this paper also applies the full-size verification approach $h(\cdot, \cdot)$~\cite{Zhao2021} to implement the classification task, \ie, outlier rejection. 
The essential matrix $\bm{\widehat{E}}$ and the putative set $\bm{C}$ are taken as inputs to produce the predicted symmetric epipolar distance set, 
\begin{equation} \label{eq4}
    \left\{ \hat{d_i} \right\} = h(\bm{\widehat{E}, \bm{C}}), \ {\forall}i \in \left\{ 1,...,N \right\}. 
\end{equation}
Finally, an empirical threshold of epipolar distance is used criterion to discriminate outliers from inliers. 

\subsection{Correspondence Masked Autoencoder} 
\label{sec:corrmae}
Inspired by the success of masked autoencoders in image recognition~\cite{He2022mae} and 3D object understanding~\cite{pang2022masked}, we focus on correspondence pruning and develop a pre-training framework that produces robust geometry-consistent representations using the encoder. 
As illustrated in Figure~\ref{fig:corrmae}, our framework is divided into three phases to perform masked inlier reconstruction, \ie, correspondence masking, correspondence learning, and masked keypoint reconstruction for $\mathcal{A}$ and $\mathcal{B}$. 
In the following, we will elaborate on the design of each phase. 

\mparagraph{Correspondence Masking}
We randomly mask correspondences with a masking ratio $r$. 
The set of masked correspondences is denoted as $\bm{S}_{gt} = ( \bm{x}_{gt}, \bm{y}_{gt} ) \in \mathbb{R}^{r{M} \times 4}$, and is used as the ground-truth for self-supervision. 
We further investigate the effects of masking ratios ranging from $40$\% to $80$\% and different masking types (random masking and block masking~\cite{yu2022point}) on method performance, see Section~\ref{sec:pretraining}. 

\mparagraph{Correspondence Learning}
The core mission of the correspondence encoder is to embed both local and global consensus for each correspondence. 
After correspondence masking, the remaining visible correspondences $\bm{S}_{vis} = ( \bm{x}_{vis}, \bm{y}_{vis} ) \in \mathbb{R}^{(1-r)M \times 4}$ are sequentially fed into the projection and encoder $\mathcal{E}_\theta$ to produce high-dimensional embeddings, referred to as visible tokens, 
\begin{gather} \label{eq5}
   \bm{T}_{vis} = \mathcal{E}_\theta(\sigma(\mathrm{MLP} (\bm{S}_{vis}))), 
\end{gather}
where $\sigma$ consists of $\mathrm{BatchNorm}$ and $\mathrm{ReLU}$ layers, and together with an $\mathrm{MLP}$, forms our projection layer. 
These tokens $\bm{T}_{vis} \in \mathbb{R}^{(1-r)M \times C}$ guide the reconstruction of the masked keypoints. 
This phase is flexible and can be replaced by any other correspondence encoder, such as recent SoTA methods~\cite{miao2024bclnet, liu2023ncm, Zhang2019, Zhao2021} or our encoder described in Section~\ref{sec:encoder}. 

\mparagraph{Masked Keypoint Reconstruction}
Due to their unordered and irregular nature, correspondences lack well-defined positional information for encoding, thereby hindering the reconstruction process. 
As illustrated in Figure~\ref{fig:corrmae}, we adopt a dual-branch structure to separately recover masked keypoints for $\mathcal{A}$ and $\mathcal{B}$, thus indirectly reconstructing the masked correspondences. 
In the $\mathcal{A}$ branch, we begin by randomly generating keypoints $\bm{x}_{mask} \in \mathbb{R}^{r{M} \times 2}$. 
The ground-truth keypoints $\bm{y}_{gt}$ from the $\mathcal{B}$ are used as positional prompts. 
These prompts are concatenated with the randomly generated keypoints and encoded into masked tokens $\bm{T}_{mask}^{\mathcal{A}} \in \mathbb{R}^{r{M} \times C}$ via an MLP. 
This simple yet crucial positional encoding manner addresses a fundamental challenge, \ie, how to reconstruct unordered correspondences. 
Next, following the vision MAE~\cite{He2022mae}, masked tokens are added to the decoder’s $\mathcal{D}_\theta$ input sequence and subsequently used to reconstruct the masked keypoints with a simple prediction head. 
Meanwhile, the decoder stacks some linear transformers~\cite{wu2022flowformer}, with fewer layers than the encoder. The above process is formulated as, 
\begin{gather} \label{eq6}
    \bm{T}_{mask}^{\mathcal{A}} =\mathrm{MLP} ( \left[ \bm{x}_{mask}, \bm{y}_{gt} \right]), \\
    \left\{ \bm{F}^{\mathcal{A}}_{i} \right\} = \mathcal{D}_\theta (  \bm{T}_{mask}^{\mathcal{A}} || \bm{T}_{vis} ), \ {\forall}i \in \left\{ mask, visible \right\}, \\
    \bm{\widehat{x}} = \mathrm{Head} (\left\{ \bm{F}^{\mathcal{A}}_{j} \right\}), \ {\forall}j \in \left\{ mask \right\}, 
\end{gather}
where $\left[ \cdot,\cdot \right]$ and $||$ denote concatenation along the channel dimension and the spatial dimension, respectively; 
$\left\{ \bm{F}^{\mathcal{A}}_{j} \right\} \in \mathbb{R}^{r {M} \times C}$ represents the decoded masked tokens from the $\mathcal{A}$ branch; 
$\mathrm{Head}$, essentially an MLP, is a prediction head; $\bm{\widehat{x}} \in \mathbb{R}^{r {M} \times 2}$ represents the reconstructed masked keypoints for the $\mathcal{A}$ branch. 
Similarly, the $\mathcal{B}$ branch performs the same operations, sharing weights for both the positional encoder and the decoder. 
Finally, the reconstructed correspondences $\bm{\widehat{S}}_{mask} = ( \bm{\widehat{x}}, \bm{\widehat{y}} ) \in \mathbb{R}^{r{M} \times 4}$ are obtained. 

\begin{figure}[t]
\begin{center}
    \includegraphics[width=0.8\linewidth]{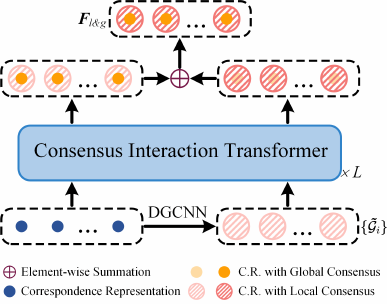}
    \vspace{-0.2cm}
    \caption{Architecture of our proposed CorrFormer. 
    C.R. is short for the correspondence representation.}\label{fig:corrformer}
\end{center}
\vspace{-0.5cm}
\end{figure}

\subsection{Correspondence Encoder}
\label{sec:encoder}
It is evident that the choice of the correspondence encoder has become a critical bottleneck in improving the capability of our method. 
To address this issue, we revisit existing encoder designs and draw the following insights. 
\textbf{First}, existing approaches typically adopt either a local-to-global~\cite{Zhao2021, liu2023ncm, dai2024mgnet, liao2023vsformer} or a bilateral learning~\cite{miao2024bclnet} paradigms. 
However, due to the lack of effective consensus interaction, they fail to effectively capture multi-scale consensus dependencies. 
\textbf{Second}, simpler structures better facilitate learning generalizable knowledge during pre-training, whereas overly complex designs~\cite{liu2023ncm} may hinder model generalization. 
\textbf{Third}, we suggest that extensible encoders are essential stepping stones toward a unified architecture. 
Motivated by these insights and by advances in multimodal encoder designs~\cite{kim2021vilt, bao2022vlmo}, we propose \textit{CorrFormer}, a simple dual-stream encoder with consensus interaction to support both pre-training and fine-tuning. 
In the following, we elaborate on the encoder during fine-tuning. 
The pre-training procedure is similar to that, except that the input switches from putative correspondences to ground-truth/generated correspondences. 

To be specific, after passing through the same projection layer as in the pre-training stage (see Section~\ref{sec:corrmae}), correspondence representations and DGCNN-encoded representations~\cite{wang2019dynamic} are separately fed into the global and local streams, as shown in Figure~\ref{fig:corrformer}. 
In DGCNN, to capture local consensus and embed it into each correspondence, we first construct $k$-NN graphs $\left\{ \bm{\mathcal{G}}_i \right\}$ (\ie, local graphs) for each correspondence based on the Euclidean distances between correspondence representations. 
The directed edge representation for each local graph is defined as, 
\begin{gather} \label{eq7}
    \bm{e_{ij}} = \left[ \bm{f_i}, \bm{f_i} - \bm{f_{ij}} \right],  \ {\forall}i \in \left\{ 1,...,N \right\}, j \in \left\{ 1,...,k \right\}, 
\end{gather}
where $\bm{f_i}$ and $\bm{f_{ij}}$ represent the $i$-th correspondence representation and its $j$-th neighbor representation, respectively. 
Furthermore, to enhance the local graph representations, a simple yet effective graph attention mechanism~\cite{liao2023vsformer}, $\mathrm{GAttn}$, is introduced into DGCNN. 
Subsequently, neighbor information for each correspondence is aggregated via annular convolution~\cite{Zhao2021}, $\mathrm{AConv}$, enabling the embedding of local consensus. 
This process can be succinctly expressed as, 
\begin{gather} \label{eq8}
    \left\{ \tilde{\bm{\mathcal{G}}_i} \right\} = \mathrm{AConv} ( \mathrm{GAttn} (\left\{ \bm{\mathcal{G}}_i \right\}) ), \ {\forall}i \in \left\{ 1,...,N \right\}.
\end{gather} 

Afterwards, the two streams undergo intensive implicit interaction through $L$ weight-shared linear transformers~\cite{wu2022flowformer}, and are fused via an element-wise summation to produce the strengthened consensus representations. 
Our CorrFormer and extra projection layer can be formulated as, 
\begin{gather} \label{eq9}
    \bm{F}_{l \& g} = \underbrace{\mathrm{LTR} ( \mathrm{Proj} (\bm{C}) )}_{\text{Global Stream}}
     + \underbrace{\mathrm{LTR} (\mathrm{DGCNN} (\mathrm{Proj} (\bm{C})))}_{\text{Local Stream}}, 
\end{gather}
where $\mathrm{Proj}$ denotes a projection layer; $\mathrm{DGCNN}$ is as described above; and $\mathrm{LTR}$ represents multiple linear transformers. 
Meanwhile, our CorrFormer’s capacity can be easily scaled up by increasing the number of intermediate transformer layers. 

\subsection{Loss Function}
\label{sec:loss}

\mparagraph{Pre-training Loss}
The overall framework is optimized using a hybrid loss function, comprising two reconstruction losses and a keypoint distribution loss, 
\begin{gather} \label{eq10}
    \mathcal{L}_\mathcal{P} = \mathcal{L}_{rec}^{\mathcal{A}} + \mathcal{L}_{rec}^{\mathcal{B}} + \lambda \mathcal{L}_{dist}, 
\end{gather}
where $\lambda$ denotes a hyperparameter that balances the two objectives. For the reconstruction objective, we use an $\ell_2$-loss between the reconstructed keypoints ($\bm{\widehat{x}}$ and $\bm{\widehat{y}}$) and the masked keypoints ($\bm{x}_{gt}$ and $\bm{y}_{gt}$), 
\begin{gather} \label{eq11}
    \mathcal{L}_{rec}^{\mathcal{A} \& \mathcal{B}} = {\| \bm{\widehat{x}} - \bm{x}_{gt} \|}_{2}^{2} + {\| \bm{\widehat{y}} - \bm{y}_{gt} \|}_{2}^{2}. 
\end{gather}

\begin{figure}[t]
\begin{center}
    \includegraphics[width=0.8\linewidth]{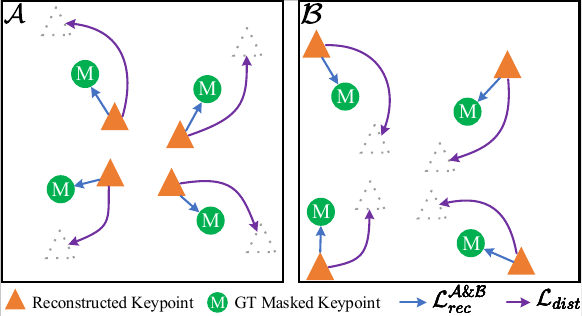}
    \caption{Functionality of the reconstruction loss and the proposed keypoint distribution loss. 
    The keypoint distribution loss aligns the relational structure of reconstructed keypoints between branches using undirected complete graphs.}\label{fig:loss}
\end{center}
\vspace{-0.5cm}
\end{figure}

A plain per-branch keypoint reconstruction loss does not explicitly enforce correspondence-level geometric consistency between the two images. 
As illustrated in Figure~\ref{fig:loss}, we therefore introduce a keypoint distribution loss to align the relational structure of keypoints across branches and encourage geometry-consistent representations. 
We first construct two undirected complete graphs based on the reconstructed keypoints from the $\mathcal{A}$ and $\mathcal{B}$ branches ($\widehat{\bm{\mathcal{G}}}^\mathcal{A}$ and $\widehat{\bm{\mathcal{G}}}^\mathcal{B}$), assigning Euclidean distances to the edges. Then, an $\ell_1$-loss is computed between the corresponding edge sets of the two graphs. The distribution loss is defined as, 
\begin{gather} \label{eq12}
    \widehat{\bm{\mathcal{G}}}^{\xi} = \left\{ e_{ij}^{\xi}|i=1,...,r{M}, j=1,...,r{M} \right\}, \\
    \mathcal{L}_{dist} = {\| \widehat{\bm{\mathcal{G}}}^\mathcal{A} - \widehat{\bm{\mathcal{G}}}^\mathcal{B} \|}_{1}, 
\end{gather}
where $\xi$ denotes either the $\mathcal{A}$ branch or the $\mathcal{B}$ branch, with $\xi \in \left\{ \mathcal{A}, \mathcal{B} \right\}$; the term $e_{ij}^{\xi}$ represents the Euclidean distance between keypoints $i$ and $j$, specifically, $e_{ij}^{\mathcal{A}} = {\| \bm{\widehat{x}}_{i} - \bm{\widehat{x}}_{j} \|}_{2}$ for the $\mathcal{A}$ branch, $e_{ij}^{\mathcal{B}} = {\| \bm{\widehat{y}}_{i} - \bm{\widehat{y}}_{j} \|}_{2}$ for the $\mathcal{B}$ branch. 

\mparagraph{Fine-tuning Loss}
We employ a commonly used binary cross-entropy loss for the binary classification problem and a geometric loss for the essential matrix regression problem, 
\begin{gather} \label{eq13}
  \mathcal{L}_\mathcal{F} = \mathcal{L}_{cls} + \beta \mathcal{L}_{ess}, 
\end{gather}
where the hyperparameter $\beta$ balances the two loss terms. Following~\cite{Zhao2021}, the classification loss $\mathcal{L}_{cls}$ is formulated as, 
\begin{equation} \label{eq14}
    \mathcal{L}_{cls} = \sum_{i=1}^{\pi} H(\bm{{\mu}}_i \odot \bm{o}_{i}, \bm{l}_i), 
\end{equation}
where $\pi$ is the number of iterations in the pruning networks; 
$H(\cdot)$ denotes a binary cross entropy loss function; 
$\bm{o}$ represents the candidate weights output by the pruning iteration network; 
$\bm{l}$ represents the weakly supervised labels, which are chosen under the symmetric epipolar distance threshold $10^{-4}$ as positive~\cite{Hartley2003}; 
$\bm{{\mu}}$ is an adaptive temperature vector; and $\odot$ represents the Hadamard product. 
Following~\cite{Zhang2019}, the geometric loss $\mathcal{L}_{ess}$ is formulated as, 
\begin{equation} \label{eq15}
    \mathcal{L}_{ess} = \frac{(\bm{p^{\prime T} \widehat{E} p})^2} {{\left\lVert \bm{{Ep}} \right\rVert}_{[1]}^{2} + {\left\lVert \bm{{Ep}} \right\rVert}_{[2]}^{2} + {\left\lVert {\bm{Ep^{\prime}}} \right\rVert}_{[1]}^{2} + {\left\lVert {\bm{Ep^{\prime}}} \right\rVert}_{[2]}^{2}},
\end{equation}
where $\bm{E}$ is the ground-truth of the essential matrix, $\widehat{\bm{E}}$ is the prediction of the last pruning network; 
$\bm{p}$ and $\bm{p^{\prime}}$ represent virtual correspondence coordinates obtained by the $\bm{E}$; 
${\left\lVert{\cdot}\right\rVert}_{[i]}$ denotes the $i$-th element of vector. 

\section{Experiments}

\subsection{Implementation Details}
\label{sec:details}

\mparagraph{Training Data}
Following the OANet~\cite{Zhang2019} and CLNet~\cite{Zhao2021} protocols, we adopt YFCC100M~\cite{Thomee2016} as the outdoor benchmark and SUN3D~\cite{Xiao2013} as the indoor benchmark (see Section~\ref{sec:pose} for details). 
Except for pre-training, all recent state-of-the-art pruners and our method are trained/fine-tuned on these benchmarks. 
Additionally, we divide GeneralPruner into two variants based on the data used for pre-training. 
In the first variant ($\dagger$ or $\ddagger$), GeneralPruner is pre-trained solely on the target benchmark, introducing no additional data. 
In the second variant ($\star$), GeneralPruner is pre-trained on an additional self-curated dataset, DiverseCorr (see Section~\ref{sec:general}), which enables us to investigate the scalability of generalization. 

\mparagraph{Pre-training Setting}
The numbers of feature channels, local graph neighbors, and linear transformer layers are set to $C=128$, $k=9$, and $L=4$, respectively. 
We adopt the AdamW optimizer~\cite{loshchilov2017decoupled} with cosine learning rate decay~\cite{loshchilov2016sgdr}. 
The initial learning rate is set to $10^{-3}$ with a weight decay of $0.05$. 
We pre-train correspondence encoders (our CorrFormer or others~\cite{Zhang2019, Zhao2021, liu2023ncm,miao2024bclnet}) for $100$ epochs with a batch size of $64$. 
The hyper-parameter $\lambda$ in Equation~\ref{eq10} is set as $0.1$. 
All pre-training experiments are conducted on multiple NVIDIA Tesla V100 GPUs. 

\mparagraph{Fine-tuning Setting}
Considering computational efficiency, we follow the CLNet~\cite{Zhao2021} protocol and adopt the SIFT detector~\cite{Lowe2004} with a nearest-neighbor matching strategy as the matcher. 
The numbers of feature channels, putative correspondences, local graph neighbors, linear transformer layers, and pruning networks are set to $C=128$, $N=2000$, $k=9$, $L=4$, and $\pi=2$, respectively. 
We employ Adam optimizer~\cite{kingma2014adam} with a weight decay of $0$ to fine-tune the pruners, and the canonical learning rate (for batch size is $32$) is set to $10^{-3}$. 
The weight $\beta$ in Equation~\ref{eq13} is set as $0$ during the first $20k$ iterations and $0.5$ in the remaining $480k$ iterations. 
All fine-tuning experiments are performed on multiple NVIDIA Tesla V100 GPUs. 

\begin{table}[t]
\caption{Quantitative comparison results (higher is better) of correspondence pruning on the outdoor benchmark across three pruner categories.}
\setlength{\tabcolsep}{3pt}
\renewcommand\arraystretch{1.}
\centering
\vspace{-0.2cm}
\begin{threeparttable}
\scalebox{1}{
\begin{tabular}{l|p{24pt}<{\centering}p{24pt}<{\centering}p{24pt}<{\centering}|p{24pt}<{\centering}p{24pt}<{\centering}p{24pt}<{\centering}}
\toprule
\multirow{2}{*}{\textbf{Method}} & \multicolumn{3}{c|}{\textbf{Pose Estimation (AUC)}} & \multicolumn{3}{c}{\textbf{Outlier Rejection ($\%$)}}
\\
& \textbf{@5\textdegree} & \textbf{@10\textdegree} & \textbf{@20\textdegree} & \textbf{Pre.} & \textbf{Rec.} & \textbf{F1}
\\\midrule\midrule
RANSAC~\cite{Fischler1981} & $3.47$ & $9.10$ & $18.60$ & $41.83$ & $57.08$ & $48.28$ 
\\
GMS~\cite{Bian2017} & $13.29$ & $24.38$ & $37.83$ & $47.75$ & $47.92$ & $47.83$ 
\\
LPM~\cite{Ma2019} & $15.99$ & $28.25$ & $41.76$ & $43.75$ & $65.65$ & $51.72$ 
\\\midrule\midrule
OANet~\cite{Zhang2019}     & $15.92$ & $35.93$ & $57.11$ & $68.05$ & $68.41$ & $68.23$ 
\\
CLNet~\cite{Zhao2021}       & $25.28$ & $45.82$ & $65.44$ & $75.05$ & $76.41$ & $75.72$ 
\\
TNet~\cite{zhong2021t}      & $20.53$ & $42.65$ & $63.20$ & $71.09$ & $72.58$ & $71.83$ 
\\
MS$^2$DGNet~\cite{Dai2022}  & $20.61$ & $42.90$ & $64.26$ & $72.61$ & $73.86$ & $73.23$ 
\\
PGFNet~\cite{liu2023pgfnet} & $24.12$ & $45.96$ & $65.06$ & $71.55$ & $72.71$ & $72.13$ 
\\
ConvMatch~\cite{convmatch2023}  & $26.83$ & $49.14$ & $67.91$ & $73.12$ & $74.39$ & $73.75$ 
\\
NCMNet~\cite{liu2023ncm}    & $34.51$ & $55.34$ & $72.40$ & $77.24$ & $78.57$ & $77.90$ 
\\
UMatch~\cite{li2024u}      & $30.84$ & $52.04$ & $69.65$ & $73.97$ & $75.72$ & $74.83$ 
\\
MGNet~\cite{dai2024mgnet}  & $32.32$ & $53.40$ & $71.59$ & $76.97$ & $79.35$ & $78.14$ 
\\
BCLNet~\cite{miao2024bclnet}   & \underline{$35.70$} & \underline{$56.62$} & \underline{$73.14$} & \underline{$77.40$} & \underline{$79.82$} & \underline{$78.59$}
\\
DeMatch~\cite{zhang2024dematch}   & $30.91$ & $52.71$ & $70.35$ & $74.11$ & $76.04$ & $75.06$ 
\\\midrule\midrule
GeneralPruner$^{\dagger}$ & $\mathbf{39.54}$ & $\mathbf{60.03}$ & $\mathbf{75.69}$ & $\mathbf{78.91}$ & $\mathbf{81.26}$ & $\mathbf{80.07}$ 
\\
\bottomrule
\end{tabular}}
\end{threeparttable}
\caption*{\scriptsize All SoTAs use official outdoor models for evaluation. 
Our GeneralPruner are pre-trained and fine-tuned solely on the outdoor benchmark~\cite{Thomee2016} without introducing any extra data. 
We \textbf{bold 1st-place} results and \underline{underline 2nd-place} results.}
\label{tab:outdoor}
\vspace{-0.6cm}
\end{table}

\subsection{Downstream Tasks}
\label{sec:tasks}

In this section, we evaluate our method alongside recent state-of-the-art pruners (SoTAs) on five downstream tasks. 
To the best of our knowledge, this is the first work to introduce a geometry-consistent pre-training method (\ie, masked inlier reconstruction) specifically designed for correspondence pruning. 
The proposed pre-training framework (\textit{CorrMAE}) and the correspondence encoder (\textit{CorrFormer}) form the core of our approach, and we jointly employ them to demonstrate the full performance of our method, referred to as GeneralPruner. 
That is, we compare our encoder equipped with pre-training to recent SoTAs with official implementations. 
For fair comparison, all pre-trained models in this section are trained solely on the target benchmarks ($\dagger$ or $\ddagger$) without involving any additional data. 

\begin{figure*}[t]
\begin{center}
    \includegraphics[width=\linewidth]{figures/vis_img_or.pdf}
    \caption{Partial typical visualization results on outdoor~\cite{Thomee2016} and indoor~\cite{Xiao2013} benchmarks. 
    The correspondence is drawn in \textcolor{green}{green} if it represents the true-positive and \textcolor{red}{red} for the false-positive.}\label{fig:or}
\end{center}
\vspace{-0.4cm}
\end{figure*}

\subsubsection{Correspondence Pruning}
\label{sec:pose}
In this paper, correspondence pruning is formulated as two complementary sub-tasks, \ie, outlier rejection and two-view pose estimation. 
Outlier rejection focuses on pruning unreliable correspondences and retaining geometrically consistent ones between image pairs to ensure robust geometric estimation. 
Two-view pose estimation aims to recover the relative camera pose, including rotation and translation, between two images of the same scene captured from different viewpoints. 
Specifically, by establishing geometrically consistent correspondences between the two views, algorithms such as direct linear transformation (DLT, implemented in this paper as a weighted eight-point algorithm~\cite{Yi2018}) or RANSAC~\cite{Fischler1981} estimate how the cameras are positioned and oriented with respect to each other. 
This task serves as a fundamental component that bridges image-level perception and geometric reasoning, forming the basis for some downstream 3D vision applications. 

\begin{table}[t]
\caption{Quantitative comparison results (higher is better) of correspondence pruning on the indoor benchmark.}
\setlength{\tabcolsep}{3pt}
\renewcommand\arraystretch{1.}
\centering
\vspace{-0.2cm}
\begin{threeparttable}
\scalebox{1}{
\begin{tabular}{l|p{24pt}<{\centering}p{24pt}<{\centering}p{24pt}<{\centering}|p{24pt}<{\centering}p{24pt}<{\centering}p{24pt}<{\centering}}
\toprule
\multirow{2}{*}{\textbf{Method}} & \multicolumn{3}{c|}{\textbf{Pose Estimation (AUC)}} & \multicolumn{3}{c}{\textbf{Outlier Rejection ($\%$)}} 
\\
& \textbf{@5\textdegree} & \textbf{@10\textdegree} & \textbf{@20\textdegree} & \textbf{Pre.} & \textbf{Rec.} & \textbf{F1} 
\\\midrule\midrule
RANSAC~\cite{Fischler1981} & $1.04$ & $3.43$ & $8.75$ & $44.11$ & $46.42$ & $45.24$ 
\\
GMS~\cite{Bian2017}  & $4.12$ & $10.53$ & $20.82$ & $41.84$ & $47.91$ & $44.67$ 
\\\midrule\midrule
OANet~\cite{Zhang2019}  & $5.93$ & $16.91$ & $34.32$ & $57.66$ & $63.10$ & $60.26$ 
\\
CLNet~\cite{Zhao2021}    & $5.85$ & $16.38$ & $32.95$ & $60.01$ & $68.09$ & $63.80$ 
\\
MS$^2$DGNet~\cite{Dai2022}  & $5.88$ & $16.83$ & $34.28$ & $57.64$ & $61.72$ & $59.61$ 
\\
NCMNet~\cite{liu2023ncm}    & \underline{$8.07$} & $20.39$ & $38.16$ & \underline{$61.11$} & $68.92$ & \underline{$64.78$} 
\\
UMatch~\cite{li2024u}    & $8.04$ & \underline{$20.83$} & \underline{$38.71$} & $58.94$ & $64.79$ & $61.73$ 
\\
BCLNet~\cite{miao2024bclnet}   & $7.68$ & $19.75$ & $37.39$ & $60.76$ & \underline{$69.08$} & $64.66$ 
\\\midrule\midrule
GeneralPruner$^{\ddagger}$  & $\mathbf{8.82}$ & $\mathbf{21.49}$ & $\mathbf{39.20}$  & $\mathbf{61.46}$ & $\mathbf{69.29}$ & $\mathbf{65.14}$ 
\\
\bottomrule
\end{tabular}}
\end{threeparttable}
\label{tab:indoor}
\vspace{-0.3cm}
\end{table}

\mparagraph{Datasets}
We evaluate all methods on both outdoor and indoor benchmarks following prior works~\cite{Yi2018, Zhang2019}. 
For the outdoor benchmark, we use the YFCC100M dataset~\cite{Thomee2016} containing $72$ reconstructed tourist landmarks ($72$ scenes)~\cite{heinly2015reconstructing}. 
Four landmarks are held out as unseen scenes for evaluation, and the remaining ones are used for training. 
After applying a visibility threshold, we obtain $540k$ image pairs for training and $4k$ for evaluation. 
Similarly, the SUN3D dataset~\cite{Xiao2013} is used as the indoor benchmark, with $239$ training scenes and $15$ unseen scenes for evaluation~\cite{ummenhofer2017demon}. 
The benchmark includes $960k$ training pairs and $15k$ evaluation pairs. 
It is worth noting that indoor scenes are more challenging due to textureless regions, repetitive structures, and self-occlusions. 

\mparagraph{Protocols}
Following previous protocols~\cite{Zhang2019, Zhao2021}, we use an efficient and lightweight matcher, \ie, SIFT~\cite{Lowe2004} with nearest-neighbor matching, to obtain $2000$ putative correspondences for each image pair. 
To ensure a fair comparison, all methods employ the full-size verification~\cite{Zhao2021} with an epipolar distance threshold of $1\times10^{-4}$ for outlier rejection evaluation. 
We report precision, recall, and F-score on both outdoor and indoor benchmarks to assess the quality of predicted correspondences. 
Based on the predicted results, we compute the essential matrix using the weighted eight-point algorithm~\cite{Yi2018}, and recover the relative camera poses from it with the OpenCV implementation. 
The pose error is measured as the maximum angular deviation in rotation and translation, and we report the area under the curve (AUC) at thresholds of $5$\textdegree, $10$\textdegree, and $20$\textdegree. 
For overall performance, we aggregate all samples from unseen scenes to obtain the mean AUC, while cross-scene generalization is detailed in Section~\ref{sec:general}. 

\mparagraph{Correspondence Results}
As shown in Tables~\ref{tab:outdoor} and~\ref{tab:indoor}, our GeneralPruner consistently outperforms all recent pruners across all evaluation metrics. 
Specifically, compared with the recent SoTA methods BCLNet~\cite{miao2024bclnet} and MGNet~\cite{dai2024mgnet}, our approach achieves improvements of $1.88$\% and $2.47$\% on the outdoor benchmark in terms of F1 score, respectively. 
On the indoor benchmark, our model also attains the best performance among all publicly available indoor pruner models. 
These results demonstrate that our method, empowered by geometry-consistent pre-training and the newly designed correspondence encoder, is able to identify more reliable correspondences between two views than previous approaches. 
Additionally, as shown in Figure~\ref{fig:or}, representative visualization results of CLNet~\cite{Zhao2021}, NCMNet~\cite{liu2023ncm}, and our method are presented from left to right. 
It can be observed that our method produces the highest correspondence quality under both outdoor and indoor scenes. 

\begin{table}[t]
\caption{Comparison of different pruners with and without our CorrMAE on the outdoor benchmark.}
\setlength{\tabcolsep}{3pt}
\renewcommand\arraystretch{1.}
\centering
\vspace{-0.2cm}
\begin{threeparttable}
\scalebox{0.9}{
\begin{tabular}{l|p{55pt}<{\centering}p{55pt}<{\centering}p{55pt}<{\centering}}
\toprule
\multirow{2}{*}{\textbf{Method}}  & \multicolumn{3}{c}{\textbf{Pose Estimation}}
\\
& \textbf{AUC@5\textdegree} & \textbf{AUC@10\textdegree} & \textbf{AUC@20\textdegree} 
\\\midrule\midrule
OANet~\cite{Zhang2019}     & $15.92$ & $35.93$ & $57.11$
\\
\myhookarrow w. CorrMAE$^{\dagger}$  & $17.40$ (\footnotesize{{$\textcolor[RGB]{125,119,34}{\mathbf{\uparrow 1.48}}$}}) & $37.55$ (\footnotesize{{$\textcolor[RGB]{125,119,34}{\mathbf{\uparrow 1.62}}$}}) & $58.34$ (\footnotesize{{$\textcolor[RGB]{125,119,34}{\mathbf{\uparrow 1.23}}$}})
\\\midrule\midrule
CLNet~\cite{Zhao2021}      & $25.28$ & $45.82$ & $65.44$
\\
\myhookarrow w. CorrMAE$^{\dagger}$  & $28.56$ (\footnotesize{{$\textcolor[RGB]{125,119,34}{\mathbf{\uparrow 3.28}}$}}) & $48.99$ (\footnotesize{{$\textcolor[RGB]{125,119,34}{\mathbf{\uparrow 3.17}}$}}) & $67.31$ (\footnotesize{{$\textcolor[RGB]{125,119,34}{\mathbf{\uparrow 1.87}}$}})
\\\midrule\midrule
NCMNet~\cite{liu2023ncm}    & $34.51$ & $55.34$ & $72.40$ 
\\
\myhookarrow w. CorrMAE$^{\dagger}$  & $35.49$ (\footnotesize{{$\textcolor[RGB]{125,119,34}{\mathbf{\uparrow 0.98}}$}}) & $56.52$ (\footnotesize{{$\textcolor[RGB]{125,119,34}{\mathbf{\uparrow 1.18}}$}}) & $73.50$ (\footnotesize{{$\textcolor[RGB]{125,119,34}{\mathbf{\uparrow 1.10}}$}}) 
\\\midrule\midrule
BCLNet~\cite{miao2024bclnet}   & $35.70$ & $56.62$ & $73.14$
\\
\myhookarrow w. CorrMAE$^{\dagger}$  & $37.88$ (\footnotesize{{$\textcolor[RGB]{125,119,34}{\mathbf{\uparrow 2.18}}$}}) & $58.61$ (\footnotesize{{$\textcolor[RGB]{125,119,34}{\mathbf{\uparrow 1.99}}$}}) & $74.91$ (\footnotesize{{$\textcolor[RGB]{125,119,34}{\mathbf{\uparrow 1.77}}$}})
\\\midrule\midrule
CorrFormer (Ours)   & $37.69$ & $58.37$ & $74.71$
\\
\myhookarrow w. CorrMAE$^{\dagger}$  & $39.54$ (\footnotesize{{$\textcolor[RGB]{125,119,34}{\mathbf{\uparrow 1.85}}$}}) & $60.03$ (\footnotesize{{$\textcolor[RGB]{125,119,34}{\mathbf{\uparrow 1.66}}$}}) & $75.69$ (\footnotesize{{$\textcolor[RGB]{125,119,34}{\mathbf{\uparrow 0.98}}$}})
\\
\bottomrule
\end{tabular}}
\end{threeparttable}
\caption*{\scriptsize All CorrMAE-based variants are pre-trained and fine-tuned from scratch on the outdoor benchmark~\cite{Thomee2016}, instead of using the official model weights.}
\label{tab:improvement}
\vspace{-0.6cm}
\end{table}

\mparagraph{Pose Estimation Results}
As shown in Table~\ref{tab:outdoor}, our GeneralPruner achieves a new state-of-the-art on the outdoor benchmark, surpassing all previous pruners by a large margin. 
On the one hand, learning-based approaches consistently outperform traditional ones, and our model improves over LPM~\cite{Ma2019} by $23.55$ AUC@5\textdegree, highlighting the advantage of deep architectures under heavy outlier conditions. 
On the other hand, compared with the recent SoTA BCLNet~\cite{miao2024bclnet}, our method achieves a further $10.76$\% improvement at AUC@5\textdegree, owing to the proposed geometry-consistent pre-training method and the novel correspondence encoder. 
To be specific, 
\textit{1) Encoder design.} The proposed encoder (\textit{CorrFormer}) adopts a dual-stream architecture with consensus interaction, 
which produces more discriminative correspondence representations and yields a $5.57$\% relative gain (the second-last row of Table~\ref{tab:improvement}) over the previous best~\cite{miao2024bclnet} at AUC@5\textdegree. 
\textit{2) CorrMAE integration.} CorrMAE enhances various pruners via masked inlier reconstruction, resulting in more robust geometry-consistent representations and boosting OANet~\cite{Zhang2019}, CLNet~\cite{Zhao2021}, NCMNet~\cite{liu2023ncm}, BCLNet~\cite{miao2024bclnet}, and our CorrFormer by $9.30$\%, $12.97$\%, $2.84$\%, $6.12$\%, and $4.91$\%, respectively (see Table~\ref{tab:improvement}). 
This also suggests that complex encoder architectures, such as NCMNet~\cite{liu2023ncm}, tend to be less effective at learning generalizable representations during pre-training. 
In addition, as shown in Table~\ref{tab:indoor}, our approach achieves the best performance among all open-source competitors on the indoor benchmark, demonstrating stronger robustness across challenging indoor scenes. 

\subsubsection{Visual Localization}
Visual localization aims to estimate the 6-degree-of-freedom camera pose of a query image with respect to a known 3D scene model. 
The ability to localize an agent within a 3D environment underpins some applications, such as robot navigation~\cite{biswas2012depth}, autonomous driving~\cite{heng2019project}, and augmented reality~\cite{ventura2014global}. 
Typical localization approaches rely on retrieval-based pipelines to establish 2D–3D (image-to-scene) correspondences, which in turn depend on robust two-view correspondences between the query and reference images. 
Thus, we adopt visual localization as a downstream task to further evaluate correspondence pruning methods. 

\mparagraph{Datasets}
We evaluate all models on the standard Aachen Day-Night benchmarks (v1.0 and v1.1)~\cite{sattler2018benchmarking, zhang2021reference} for outdoor visual localization. 
These benchmarks cover the historic inner city of Aachen, Germany, spanning approximately $6km^2$, and is characterized by large viewpoint changes and severe day–night illumination variations. 
Aachen v1.0 contains $4328$ reference images and $922$ query images ($824$ day and $98$ night). 
Aachen v1.1 extends v1.0 with $2369$ additional reference images and $93$ additional night queries, resulting in a total of $6697$ references, $824$ day queries, and $191$ night queries. 
We follow the official reference–query splits and report results on both versions. 

\begin{table}[t]
\caption{Quantitative comparison results (higher is better) of visual localization on two benchmarks~\cite{sattler2018benchmarking, zhang2021reference}.}
\setlength{\tabcolsep}{3pt}
\renewcommand\arraystretch{1.}
\centering
\vspace{-0.2cm}
\begin{threeparttable}
\scalebox{1}{
\begin{tabular}{l|c|cc}
\toprule
\multirow{2}{*}{\textbf{Method}} & \textbf{Day} & \textbf{Night}
\\\cmidrule{2-3}
& \multicolumn{2}{c}{\textbf{($0.25m$, $2$\textdegree) / ($0.5m$, $5$\textdegree) / ($5m$, $10$\textdegree)}}
\\\cmidrule{1-3}
\multicolumn{3}{c}{Visual Localization of Aachen v1.0}
\\\midrule\midrule
NoPruner~\cite{Lowe2004}  & $82.3$ / $88.2$ / $92.7$ & $38.8$ / $45.9$ / $57.1$  
\\
OANet~\cite{Zhang2019} & \underline{$85.4$} / $92.4$ / $96.6$ & $63.3$ / $74.5$ / $83.7$	 
\\
CLNet~\cite{Zhao2021}    & $85.0$ / \underline{$93.0$} / $97.7$ & $67.3$ / $81.6$ / $88.8$	 
\\
ConvMatch~\cite{convmatch2023} &  $85.2$ / $91.9$ / $96.1$ & $58.2$ / $63.3$ / $75.5$	 
\\
NCMNet~\cite{liu2023ncm} & $85.2$ / $92.7$ / \underline{$98.1$} & \underline{$68.4$} / \underline{$81.6$} / \underline{$90.8$}
\\\midrule\midrule
GeneralPruner$^{\dagger}$ & $\mathbf{86.4}$ / $\mathbf{93.4}$ / $\mathbf{98.2}$ & $\mathbf{76.5}$ / $\mathbf{83.7}$ / $\mathbf{94.9}$ 
\\\cmidrule{1-3}
\multicolumn{3}{c}{Visual Localization of Aachen v1.1}
\\\midrule\midrule
NoPruner~\cite{Lowe2004}  & $85.4$ / $91.1$ / $94.5$ & $35.6$ / $42.4$ / $55.0$  
\\
OANet~\cite{Zhang2019} & \underline{$87.7$} / \underline{$94.8$} / \underline{$98.1$} & $58.6$ / $69.1$ / $83.8$	 
\\
CLNet~\cite{Zhao2021}    & $86.7$ / $94.1$ / $98.1$ & $61.3$ / \underline{$78.0$} / $89.0$	 
\\
ConvMatch~\cite{convmatch2023} &  $87.3$ / $93.8$ / $97.5$ & $51.8$ / $60.2$ / $73.3$	 
\\
NCMNet~\cite{liu2023ncm} &  $86.7$ / $94.1$ / $98.4$ & \underline{$61.3$} / $77.5$ / \underline{$90.1$}  
\\\midrule\midrule
GeneralPruner$^{\dagger}$ & $\mathbf{88.2}$ / $\mathbf{95.4}$ / $\mathbf{98.9}$ & $\mathbf{66.5}$ / $\mathbf{82.7}$ / $\mathbf{94.2}$
\\
\bottomrule
\end{tabular}}
\end{threeparttable}
\caption*{\scriptsize All models use their outdoor-trained versions for this evaluation.}
\label{tab:visloc}
\vspace{-0.6cm}
\end{table}

\mparagraph{Protocols}
Following ConvMatch~\cite{convmatch2023} protocols, all pruners are plugged into the official HLoc~\cite{sarlin2019coarse} pipeline, 
while SIFT~\cite{Lowe2004} and mutual nearest-neighbor matching (MNN) are leveraged to produce $4096$ putative correspondences. 
The predicted correspondences between reference and query images are obtained in the same manner as in the aforementioned correspondence pruning task. 
Subsequently, query poses are estimated using PnP~\cite{kneip2011novel} with RANSAC~\cite{Fischler1981} for robustness, and the evaluation follows the official setup\textsuperscript{\ref{fn:website}}. 
We report localization accuracy under distance and angular error thresholds of ($0.25m$, $2$\textdegree), ($0.5m$, $5$\textdegree), and ($5m$, $10$\textdegree). 

\begin{tikzpicture}[remember picture, overlay]
  \node[anchor=south east, xshift=-14.2cm, yshift=0.7cm] at (current page.south east)
  {\scriptsize\refstepcounter{footnote}\label{fn:website}\textsuperscript{\thefootnote}
   \href{https://www.visuallocalization.net/benchmark}{https://www.visuallocalization.net/benchmark}};
\end{tikzpicture}

\mparagraph{Results}
As shown in Table~\ref{tab:visloc}, our GeneralPruner achieves the best overall performance on both Aachen v1.0 and v1.1 benchmarks, demonstrating strong robustness under severe day–night illumination changes. 
All pruner-equipped methods clearly outperform the baseline without pruning (\ie, SIFT + MNN), highlighting the essential role of correspondence pruning in zero-shot visual localization. 
In particular, our approach consistently surpasses all recent pruners across localization thresholds, with especially large gains in night-time scenarios. 
For example, under the strictest threshold of ($0.25m$, $2$\textdegree), our model improves over the second-best by $11.84$\% on Aachen v1.0 ($68.4$ vs. our $76.5$) and by $8.48$\% on Aachen v1.1 ($61.3$ vs. our $66.5$). 
These results confirm the superior generalization capability of our model across different lighting conditions and benchmark versions. 
This should be attributed to the generalizable geometry-consistent representation learned through the masked inlier reconstruction. 

\subsubsection{Camera Tracking}
Camera tracking plays a pivotal role in visual SLAM systems~\cite{mur2017orb, murai2025mast3r}, directly influencing the accuracy of camera localization and the quality of online map construction in unknown environments. 
Among its key components, pose estimation provides the foundation for achieving robust and consistent tracking~\cite{li20254d}. 
In this work, we adopt the camera tracking task to further evaluate how recent correspondence pruning methods perform in the practical application. 

\begin{figure*}[t]
\begin{center}
    \includegraphics[width=\linewidth]{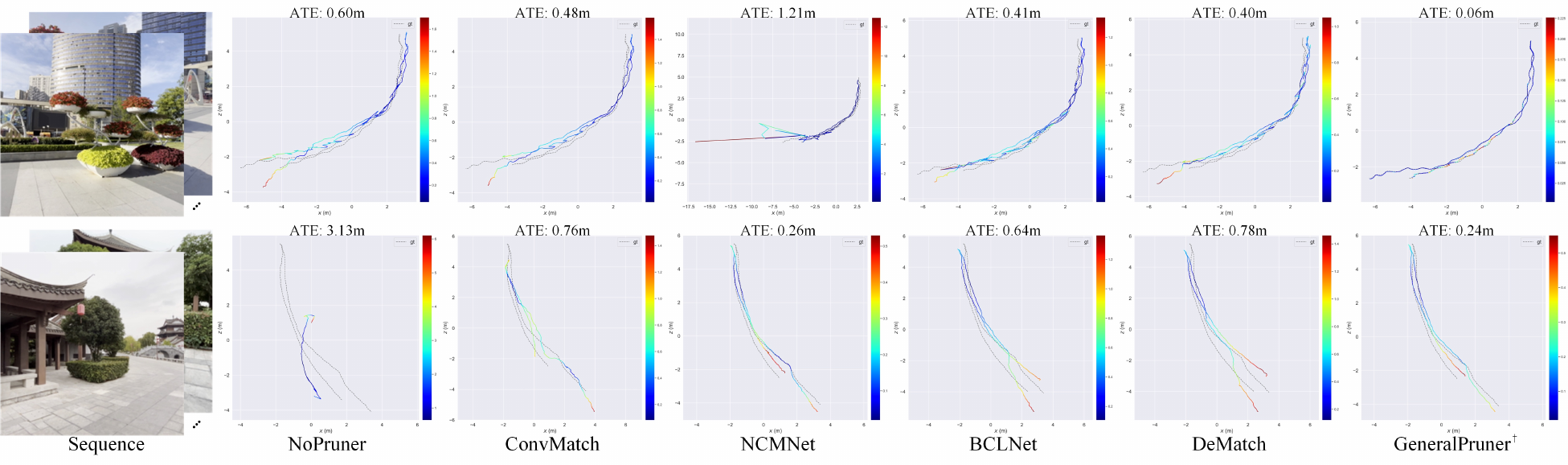}
    \caption{Visualization of camera trajectories on two outdoor sequences from the DL3DV dataset~\cite{ling2024dl3dv}. 
    The \textcolor{orange!80!red}{colored trajectories} represent the predicted ones, and \textcolor{gray}{gray dashed lines} are the ground-truth trajectorie. }\label{fig:track}
\end{center}
\vspace{-0.5cm}
\end{figure*}

\begin{table}[t]
\caption{Quantitative comparison results (lower is better) of camera tracking on three scenes from DL3DV~\cite{ling2024dl3dv}.}
\setlength{\tabcolsep}{3pt}
\renewcommand\arraystretch{1.}
\centering
\vspace{-0.2cm}
\begin{threeparttable}
\scalebox{1}{
\begin{tabular}{l|p{29pt}<{\centering}|p{29pt}<{\centering}|p{29pt}<{\centering}|p{29pt}<{\centering}}
\toprule
\multirow{2}{*}{\textbf{Method}}  & \multicolumn{4}{c}{\textbf{Camera Tracking (ATE RMSE $\downarrow$ [m])}}
\\\cmidrule{2-5}
& \textbf{Seq. 01} & \textbf{Seq. 02} & \textbf{Seq. 03} & \textbf{Avg.}
\\\midrule\midrule
NoPruner~\cite{potje2024xfeat}  & $0.50$ & $0.76$ & $3.18$ & $1.48$
\\
OANet~\cite{Zhang2019}  & \underline{$0.44$} & $0.76$ & $2.96$ & $1.39$
\\
CLNet~\cite{Zhao2021}  & $0.60$ & $2.56$ & $2.15$ & $1.77$
\\
TNet~\cite{zhong2021t}  & $0.46$ & $3.17$ & $0.84$ & $1.49$ 
\\
MS$^2$DGNet~\cite{Dai2022}  & $0.53$ & $1.11$ & $2.89$ & $1.51$ 
\\
ConvMatch~\cite{convmatch2023}   & $0.52$ & $0.58$ & $0.82$ & $0.64$
\\
NCMNet~\cite{liu2023ncm}    & $\mathbf{0.41}$ & $1.64$ & \underline{$0.33$} & $0.79$ 
\\
UMatch~\cite{li2024u}   & $0.48$ & \textcolor{gray}{$TL$} & \textcolor{gray}{$TL$} & \textcolor{gray}{$/$}
\\
MGNet~\cite{dai2024mgnet}   & $0.47$ & $2.87$ & $1.19$ & $1.51$ 
\\
BCLNet~\cite{miao2024bclnet}   & $0.57$ & \underline{$0.48$} & $0.70$ & \underline{$0.58$} 
\\
DeMatch~\cite{zhang2024dematch} & $0.67$ & $0.49$ & $0.67$ & $0.61$ 
\\\midrule\midrule 
GeneralPruner$^{\dagger}$  & $0.45$ & $\mathbf{0.16}$ & $\mathbf{0.26}$ & $\mathbf{0.29}$ 
\\
\bottomrule
\end{tabular}}
\end{threeparttable}
\caption*{\scriptsize All models use outdoor versions for this evaluation. 
\textcolor{gray}{$TL$} is short for tracking lost.
}
\label{tab:track}
\vspace{-0.6cm}
\end{table}

\mparagraph{Datasets}
We conduct experiments on the DL3DV dataset~\cite{ling2024dl3dv} to evaluate the tracking accuracy of recent pruners in outdoor environments. 
Following the setup of the recent outdoor monocular SLAM method~\cite{cheng2025outdoor}, we select the same three $300$-frame sequences from DL3DV. 
The selected scenes are static and exhibit significant camera viewpoint changes. 

\mparagraph{Protocols}
We adopt the de facto standard in real-world applications, \ie, the open-source ORB-SLAM2 system~\cite{mur2017orb}, as our evaluation toolkit to assess recent pruners. 
To better align with modern feature extraction trends, the original ORB detector~\cite{rublee2011orb} is replaced with the recent XFeat~\cite{potje2024xfeat} and paired with MNN to form an efficient matcher that generates $2000$ putative correspondences per image pair. 
In addition, a pre-trained DBoW3 vocabulary~\cite{galvez2012bags} is employed to support loop closure detection. 
Afterward, all pruners are integrated into the system as plug-in modules for unified evaluation. 
During tracking, the predicted correspondences between current and reference frames are further pruned using an epipolar distance threshold of $1\times10^{-2}$. 
Except for these modifications, all other configurations remain identical to the vanilla ORB-SLAM2 setup. 
For tracking evaluation, we report the Root Mean Square Error (RMSE) of the Absolute Trajectory Error (ATE) over all keyframes. 
To ensure evaluation stability, all experiments are conducted in single-thread mode and repeated five times, with the average result reported. 

\mparagraph{Results}
As shown in Table~\ref{tab:track}, GeneralPruner achieves the lowest average tracking error across three outdoor sequences of the DL3DV dataset, demonstrating superior camera tracking performance in realistic SLAM settings. 
Most pruner-equipped variants outperform the baseline without pruning (\ie, XFeat + MNN), indicating that pruners effectively enhance tracking stability. 
In particular, our method attains an average ATE RMSE of $0.29m$, reducing the error by a half compared to the best previous pruner ($0.58m$). 
The improvement mainly comes from \textit{Seq.02} and \textit{Seq.03}, which contain richer visual features than \textit{Seq.01}, offering greater potential for accuracy enhancement. 
These results highlight the practical value of our approach for achieving robust camera tracking. 
Additionally, as shown in Figure~\ref{fig:track}, we visualize the camera trajectories of different methods on \textit{Seq.02} and \textit{Seq.03}. 
Our trajectories align much more closely with the ground-truth, further confirming the advantage of our approach. 
We attribute these gains to the proposed geometry-consistent pre-training paradigm, which sculpts robust and generalizable correspondence representations, leading to more stable pose estimation during tracking. 

\begin{figure}[t]
\begin{center}
    \includegraphics[width=\linewidth]{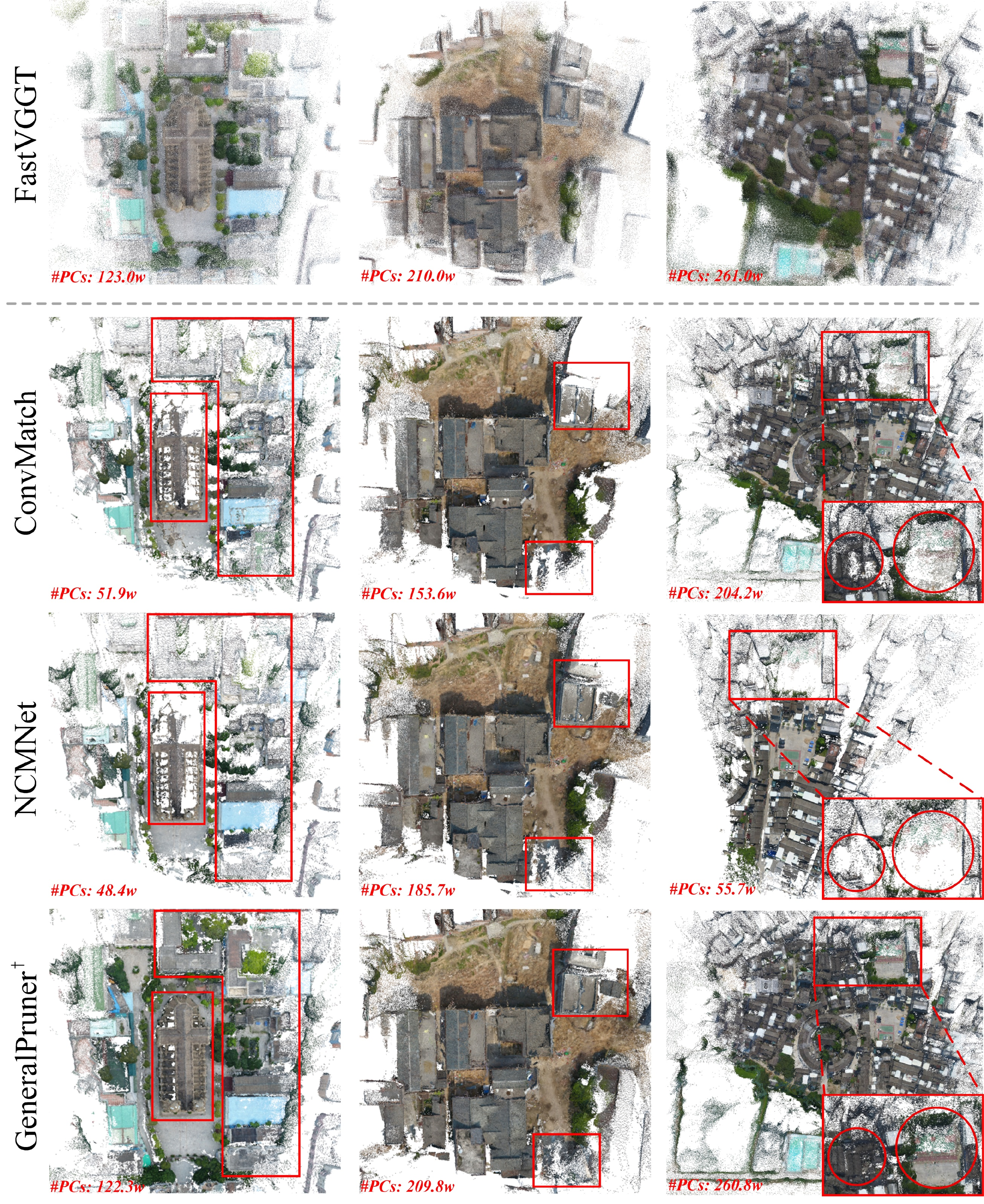}
    \vspace{-0.2cm}
    \caption{Qualitative comparison of reconstructed 3D maps on three outdoor scenes from the BlendedMVS~\cite{yao2020blendedmvs}.}\label{fig:3d_rec}
\end{center}
\vspace{-0.5cm}
\end{figure}

\subsubsection{3D Reconstruction}
3D reconstruction based on Structure-from-Motion (SfM)~\cite{agarwal2011building, schonberger2016structure} and Multi-View Stereo (MVS)~\cite{galliani2015massively, wang2023adaptive} aims to recover a 3D scene offline from a set of unconstrained multi-view images. 
As one of the most prevalent paradigms in the 3D reconstruction field, 
SfM and MVS are widely adopted in real-world applications such as archaeology, cultural heritage preservation, and medical assistance. 
Recently, end-to-end SfM methods~\cite{wang2025vggt, shen2025fastvggt} have emerged as an alternative paradigm that formulates the reconstruction pipeline in a fully neural manner. 
These methods aim to simplify classical optimization and enable fast 3D reconstruction when sufficient computational resources are available. 
Despite their promise, they typically require substantial computational and memory resources and still face challenges in reconstruction robustness for complex real-world scenes. 
As a result, classical SfM pipelines such as COLMAP~\cite{schonberger2016structure} remain widely adopted in practice due to their efficiency and robustness. 
In classical SfM pipelines, reconstruction quality largely depends on the accuracy of geometric verification. 
Correspondence pruning in the front-end can improve this process by mining geometrically consistent correspondences in advance~\cite{liu2023ncm}. 
To this end, we employ the SfM-based 3D reconstruction task to further evaluate the performance of recent state-of-the-art correspondence pruning methods. 

\mparagraph{Datasets}
We use the BlendedMVS dataset~\cite{yao2020blendedmvs} to evaluate the reconstruction quality of recent state-of-the-art pruners. 
To better highlight their capability, we focus on three representative and challenging large-scale outdoor scenes, 
namely \textit{Sacred Heart Cathedral}, \textit{Village Corner}, and \textit{Hakka Village}, rather than on small or object-centric scenes. 
These scenes consist of $124$, $270$, and $375$ aerial images captured by drones, all with a resolution of $768\times576$ pixels. 

\mparagraph{Protocols}
We adopt the de facto standard in offline 3D reconstruction, \ie, the open-source COLMAP framework, as our evaluation pipeline to assess recent pruners. 
To ensure computational efficiency, the number of extracted SIFT~\cite{Lowe2004} keypoints is restricted to $2000$ per image, followed by nearest-neighbor matching to establish putative correspondences between image pairs. 
Afterward, all pruners are integrated into the COLMAP as modular plug-ins, with predicted correspondences obtained using the same settings as in the correspondence pruning evaluation. 
Except for these modifications, all other configurations remain identical to the vanilla COLMAP setup. 
Besides, since the evaluation scenes do not exhibit the so-called \textit{big ben problem}\textsuperscript{\ref{fn:bigben}}, we did not incorporate visual disambiguation techniques~\cite{cai2023doppelgangers}. 
For reconstruction quality, we report the number of points in the reconstructed sparse 3D maps (denoted as \textit{PCs}) and visualize the results from a bird’s-eye-view perspective. 

\mparagraph{Results}
As presented in Figure~\ref{fig:3d_rec}, GeneralPruner achieves the densest and most complete 3D reconstructions on all three outdoor scenes from BlendedMVS~\cite{yao2020blendedmvs}. 
While ConvMatch~\cite{convmatch2023} and NCMNet~\cite{liu2023ncm} tend to produce incomplete or fragmented structures, 
our method yields dense and coherent reconstructions with significantly higher point cloud counts ($122.3w$, $209.8w$, and $260.8w$). 
The visual comparisons in the red-marked areas show clearer contours and more consistent geometry. 
We further compare our results with a recent end-to-end SfM system, FastVGGT~\cite{shen2025fastvggt}. 
It often produces reconstructions with noticeable noise and less stable geometry on these outdoor scenes. 
In contrast, COLMAP equipped with our correspondence pruner yields more accurate and robust reconstructions. 
This performance advantage primarily arises from the geometry-consistent pre-training, which strengthens generalization and stabilizes correspondence pruning during triangulation. 

\begin{figure*}[t]
\begin{center}
    \includegraphics[width=0.9\linewidth]{figures/vis_pcd_reg.pdf}
    \vspace{-0.2cm}
    \caption{Visualization results of outlier rejection and 3D registration on two typical examples from 3DMatch~\cite{zeng20173dmatch}.}\label{fig:reg}
\end{center}
\vspace{-0.4cm}
\end{figure*}

\begin{table}[t]
\caption{Quantitative comparison results on the 3DMatch benchmark~\cite{zeng20173dmatch} for 3D registration.}
\setlength{\tabcolsep}{3pt}
\renewcommand\arraystretch{1.}
\centering
\vspace{-0.2cm}
\begin{threeparttable}
\scalebox{0.9}{
\begin{tabular}{l|p{28pt}<{\centering}p{28pt}<{\centering}p{28pt}<{\centering}|p{28pt}<{\centering}p{28pt}<{\centering}p{28pt}<{\centering}}
\toprule
\multirow{2}{*}{\textbf{Method}} & \multicolumn{3}{c|}{\textbf{3D Registration}} & \multicolumn{3}{c}{\textbf{Outlier Rejection}} 
\\
& \textbf{RE ($\downarrow$)} & \textbf{TE ($\downarrow$)} & \textbf{RR ($\uparrow$)} & \textbf{Pre. ($\uparrow$)} & \textbf{Rec. ($\uparrow$)} & \textbf{F1 ($\uparrow$)} 
\\\midrule\midrule
CLNet~\cite{Zhao2021}   & $3.10$ & $8.88$ & $74.06$ & $56.61$ & $65.55$ & $57.98$ 
\\
PGFNet~\cite{liu2023pgfnet}   & $3.15$ & $9.14$ & $73.38$  & $58.49$ & $57.69$ & $55.12$ 
\\
ConvMatch~\cite{convmatch2023}  & $3.66$ & $10.22$ & $62.97$  & $52.17$ & $50.13$ & $47.62$ 
\\
NCMNet~\cite{liu2023ncm}   & $2.66$ & $7.84$ & $75.48$ & $62.58$ & $59.63$ & $59.22$ 
\\
MGNet~\cite{dai2024mgnet}     & $2.92$ & $8.39$ & $75.23$ & $57.76$ & $\mathbf{73.91}$ & $62.50$
\\
BCLNet~\cite{miao2024bclnet}     & \underline{$2.50$} & \underline{$7.68$} & $73.51$ & \underline{$63.60$} & $54.32$ & $56.30$ 
\\
DeMatch~\cite{zhang2024dematch}    & $3.44$ & $10.18$ & $64.02$ & $50.29$ & $45.57$ & $45.58$ 
\\
CorrFormer (ours)   & $2.91$ & $8.45$ & \underline{$78.07$} & $61.97$ & $68.95$ & \underline{$63.49$}  
\\\midrule\midrule
\myhookarrow w. CorrMAE$^{\dagger}$   & $\mathbf{2.45}$ & $\mathbf{7.54}$ & $\mathbf{82.01}$ & $\mathbf{70.20}$ & \underline{$70.44$} & $\mathbf{68.88}$  
\\
\bottomrule
\end{tabular}}
\end{threeparttable}
\label{tab:regis}
\vspace{-0.3cm}
\end{table}

\begin{tikzpicture}[remember picture, overlay]
  \node[anchor=south east, xshift=-6.4cm, yshift=0.7cm] at (current page.south east)
  {\scriptsize\refstepcounter{footnote}\label{fn:bigben}\textsuperscript{\thefootnote}
   London's Big Ben is a clock tower with four-way symmetry, where the four sides of the tower look nearly the same.~\cite{cai2023doppelgangers}};
\end{tikzpicture}

\subsubsection{3D Registration}
3D registration aims to align two partially overlapping 3D point clouds by estimating a rigid transformation (rotation and translation) between them. 
This task enhances an agent’s ability to perceive and understand the surrounding 3D environment~\cite{huang2024kdd, takmaz2023openmask3d}. 
Typical registration pipelines follow a paradigm similar to correspondence pruning, as both aim to learn and refine geometrically consistent correspondences before performing geometric estimation. 
We employ 3D registration as a downstream task to evaluate the capability of recent correspondence encoders across different geometric spaces (\ie from $\mathbb{R}^4$ to $\mathbb{R}^6$). 
In addition, we further explore the cross-space transferability of the pre-trained geometry-consistent representations. 

\mparagraph{Datasets}
We evaluate recent pruners on the 3DMatch benchmark~\cite{zeng20173dmatch}, which consists of $62$ scenes with RGB-D scans. 
Following the official split, $46$ scenes are used for training, $8$ for validation, and $8$ for testing. 
All point clouds are voxel-downsampled with a $5cm$ grid, and each point cloud pair shares more than $30$\% spatial overlap, ensuring sufficient geometric consistency for evaluation. 

\mparagraph{Protocols}
Following NCMNet~\cite{liu2023ncm} protocols, we replace the representation learning module in the standard 3D registration pipeline~\cite{zhao2025progressive} with recent correspondence encoders, 
then retrain and evaluate it on the 3DMatch benchmark~\cite{zeng20173dmatch}. 
The putative correspondences are established using learning-based descriptor FCGF~\cite{choy2019fully} combined with MNN. 
To better assess the pruners’ representation capability across geometric spaces, we adopt a simplified setting compared with NCMNet, in which the rigid transformation is estimated solely through weighted SVD without any post-processing. 
All other training and evaluation configurations remain identical to the standard 3D registration protocol~\cite{zhao2025progressive}. 
We report Registration Recall (RR), Rotation Error (RE), Translation Error (TE), and correspondence-level metrics including precision, recall, and F-score. 
RR represents the percentage of successful alignment with RE and TE below $15$\textdegree~and $30cm$, respectively. 
Notably, RR is regarded as the most important and representative metric for 3D registration, whereas RE and TE are reported as supplementary measures for precision analysis. 

\mparagraph{Results}
As shown in the second-last row of Table~\ref{tab:regis}, our proposed dual-stream encoder with consensus interaction outperforms all other correspondence encoders on both 3D registration and outlier rejection tasks, as reflected by the key metrics RR and F1. 
This demonstrates that our encoder possesses stronger cross-space representation capability compared with previous correspondence encoders. 
Furthermore, as presented in the last row of Table~\ref{tab:regis}, when equipped with the pre-trained geometry-consistent representation learned from 2D-correspondences, our model achieves a significant performance gain. 
Specifically, it surpasses the second-best approach by $5.05$\% in RR and $8.49$\% in F1. 
This is an intriguing finding, suggesting that the geometric consistency learned from 2D-correspondences can effectively benefit the learning of 3D-correspondences. 
It further confirms that our pre-trained geometry-consistent representation exhibits strong transferability across different geometric spaces. 
Finally, as shown in Figure~\ref{fig:reg}, representative visualization results of CLNet~\cite{Zhao2021}, NCMNet~\cite{liu2023ncm}, and our method are presented from left to right. 

\begin{table}[t]
\caption{Ablation studies on the outdoor benchmark for CorrFormer design choices.}
\setlength{\tabcolsep}{3pt}
\renewcommand\arraystretch{1.}
\centering
\vspace{-0.2cm}
\begin{threeparttable}
\scalebox{1.}{
\begin{tabular}{l|c@{\hskip 11pt}cc}
\toprule
\multirow{2}{*}{\textbf{Method}} & \multicolumn{3}{c}{\textbf{Pose Estimation AUC}}
\\\cmidrule{2-4}
& \textbf{@5\textdegree} & \textbf{@10\textdegree} & \textbf{@20\textdegree}
\\\midrule\midrule
CorrFormer  & $\textbf{37.69}$ & $\textbf{58.37}$ & $\textbf{74.71}$
\\
\myhookarrow (a) w/o Local Stream  & $34.64$ & $54.79$ & $71.19$ 
\\
\myhookarrow (b) w/o Global Stream  & $28.65$ & $49.33$ & $67.59$ 
\\
\myhookarrow (c) L2G Structure (w/o Interaction)  & $36.00$ & $56.27$ & $73.00$ 
\\
\myhookarrow (d) w. Cross-Attn. Interaction  & $36.92$ & $57.89$ & $74.46$ 
\\
\bottomrule
\end{tabular}}
\end{threeparttable}
\label{tab:encoder}
\vspace{-0.3cm}
\end{table}

\subsection{Ablation Study}
\label{sec:ablation}
In this section, we conduct in-depth ablation studies to fully understand two core components (\ie, CorrFormer and CorrMAE) of our method. 
We first validate the design effectiveness of the proposed correspondence encoder through a set of experiments. 
Subsequently, the most effective encoder identified in the previous study is incorporated into CorrMAE, where we further explore various pre-training strategies on the masked inlier reconstruction pretext task. 
All ablation experiments are performed on the outdoor benchmark YFCC100M~\cite{Thomee2016} for two-view pose estimation. 

\begin{table}[t]
\caption{Ablation studies on the outdoor benchmark for the pre-training strategies of CorrMAE.}
\setlength{\tabcolsep}{3pt}
\renewcommand\arraystretch{1.}
\centering
\vspace{-0.2cm}
\begin{threeparttable}
\scalebox{0.95}{
\begin{tabular}{l|c|c@{\hskip 11pt}cc}
\toprule
\multirow{2}{*}{\textbf{Method}} & \multirow{2}{*}{\textbf{Loss}} & \multicolumn{3}{c}{\textbf{Pose Estimation AUC}}
\\\cmidrule{3-5}
& & \textbf{@5\textdegree} & \textbf{@10\textdegree} & \textbf{@20\textdegree}
\\\midrule\midrule
CorrMAE$^{\dagger}$  & $1.920$ & $\textbf{39.54}$ & $\textbf{60.03}$ & $\textbf{75.69}$
\\\midrule\midrule
\myhookarrow (a) Random Masking with $40\%$   & $1.787$ & $38.84$ & $59.05$ & $74.67$ 
\\
\myhookarrow (b) Random Masking with $80\%$   & $2.209$ & $38.25$ & $58.72$ & $74.80$ 
\\
\myhookarrow (c) Block Masking with $60\%$  & $2.086$ & $38.75$ & $58.93$ & $74.63$ 
\\\midrule\midrule
\myhookarrow (d) w/o Distribution Loss  & $1.913$ & $39.07$ & $59.53$ & $75.24$ 
\\\midrule\midrule
\myhookarrow (e) Shuffled Positional Prompts  & $5.451$ & $38.02$  & $59.09$ & $75.37$
\\
\myhookarrow (f) Positional Prompts w/o mask  & $1.924$ & $39.05$  & $59.61$  & $75.45$  
\\\midrule\midrule
\myhookarrow (g) SIFT+NN (Outliers $89.23\%$)  & $4.658$ & $37.73$  & $58.81$  & $75.13$ 
\\
\myhookarrow (h) SuperGlue (Outliers $25.33\%$)  & $0.764$ & $38.01$  & $59.11$  & $75.35$
\\
\myhookarrow (i) LoFTR (Outliers $15.18\%$) & $0.578$ & $38.40$ & $59.16$  & $75.43$
\\
\myhookarrow (j) DKM (Outliers $6.10\%$)  & $0.157$ & $39.75$  & $60.22$  & $75.77$
\\
\bottomrule
\end{tabular}}
\end{threeparttable}
\label{tab:pretrain}
\vspace{-0.3cm}
\end{table}

\subsubsection{Encoder Design}

\mparagraph{Single and Dual Streams}
As shown in Table~\ref{tab:encoder}, using only the global stream by removing the local stream (Row a) forms a plain encoder with four stacked linear transformers. 
This leads to a clear performance drop of $8.09$\% at AUC@5\textdegree~compared with our dual-stream CorrFormer. 
Similarly, when keeping only the local stream (Row b), the network relies solely on the DGCNN-encoded~\cite{wang2019dynamic} local consensus for learning, causing an even larger degradation. 
These results show two key findings. 
\textit{1) First}, both local and global consensus are necessary for effective correspondence pruning. 
\textit{1) Second}, global consensus plays a more fundamental role, aligning with the principle that geometric estimation favors global optimality. 
By jointly incorporating both types of consensus, our dual-stream design achieves more accurate and robust pruning performance. 

\mparagraph{Structure}
As shown in Row (c) of Table~\ref{tab:encoder}, we modify our encoder into a widely used local-to-global structure~\cite{Zhao2021, Dai2022, liu2023ncm, dai2024mgnet, liao2023vsformer}, where the local and global consensus are directly fused through skip connections without any interaction mechanism. 
Compared with our design that incorporates consensus interaction, the performance decreases by $4.48$\%. 
This observation aligns with the intuition that different types of consensus should first interact before fusion, as such interaction enables richer and more discriminative correspondence representations. 

\mparagraph{Interaction}
As shown in Row (d) of Table~\ref{tab:encoder}, we replace our consensus interaction mechanism with cross-attention. 
Specifically, the stacked transformers alternately perform self- and cross-attention between the global and local streams. 
However, this design does not lead to further performance improvement. 
We attribute this to the fact that excessive explicit interaction may hinder representation learning for correspondence pruning. 
In contrast, our implicit interaction design proves to be more effective. 

\subsubsection{Pre-training Strategy}
\label{sec:pretraining}

\mparagraph{Masking}
As shown in Rows (a) and (b) of Table~\ref{tab:pretrain}, we first explore the suitable masking ratio for the masked inlier reconstruction pretext task. 
Rows (a), CorrMAE, and (b) correspond to random masking ratios of $40$\%, $60$\%, and $80$\%, respectively. 
After fine-tuning on the pose estimation, the $60$\% masking ratio yields the best performance, suggesting a good balance between task difficulty and information retention. 
We further investigate the effect of masking type. 
As shown in Row (c), replacing random masking with block masking (under the same $60$\% ratio) results in higher pre-training loss and lower fine-tuning accuracy. 
This indicates that random masking better encourages the model to learn generalized geometric representations. 
Therefore, CorrMAE adopts random masking with a $60$\% ratio as the default setting for supporting downstream correspondence pruning. 

\mparagraph{Distribution Loss}
As shown in Row (d) of Table~\ref{tab:pretrain}, we remove the proposed distribution loss to verify its effectiveness. 
Our method achieves a further improvement on the pose estimation. 
That is, we use the proposed keypoint distribution loss between the reconstructed keypoints of both the source and target branches, proving effective for inlier representation learning. 

\mparagraph{Positional Prompts}
We further analyze two key design choices in CorrMAE, namely the ordering of positional prompts and the use of random mask embeddings. 
We first randomly shuffle the positional prompts before feeding them into the two branches. 
This breaks the correspondence-consistent ordering and introduces misleading positional cues, which makes masked correspondence reconstruction unreliable. 
As a result, the model fails to learn stable, inlier-consistent representations and exhibits clearly degraded performance, see Row (e) of Table~\ref{tab:pretrain}. 
We then remove the mask embeddings and directly input the positional prompts without any masking noise. 
Unlike shuffling, this variant remains trainable and achieves performance close to the default setting (Row (f)). 
Nevertheless, the full model with mask embeddings consistently performs better, 
indicating that the mask/noise mechanism provides a non-trivial benefit by discouraging shortcut learning and encouraging context-based reconstruction. 

\mparagraph{Outlier Sensitivity}
We study how the outlier level in pre-training correspondences affects downstream performance. 
We generate pre-training inputs using different matchers with varying outlier ratios, ranging from high-outlier SIFT+NN~\cite{Lowe2004} ($89.23\%$) to learning-based matchers such as SuperGlue~\cite{Sarlin2020} ($25.33\%$), LoFTR~\cite{sun2021loftr} ($15.18\%$), and DKM+RANSAC~\cite{edstedt2023dkm,Fischler1981} ($6.10$\%). 
As shown in Rows (g)-(j) of Table~\ref{tab:pretrain}, high outlier ratios lead to poorer pre-training and weaker downstream results. 
In contrast, as the correspondence quality improves, the downstream pose estimation consistently improves. 
Interestingly, pre-training with DKM+RANSAC slightly outperforms using GT inliers ($39.75$ \vs $39.54$ at AUC@5\textdegree), 
suggesting that a small amount of residual outliers can act as beneficial noise and improve robustness. 

\subsection{Generalization Evaluation}
\label{sec:general}
To comprehensively investigate the generalization capability of recent correspondence pruning methods, we conduct experiments from two perspectives, \ie, cross-scene and cross-domain, on the two-view pose estimation task. 
Beforehand, we collect a dataset (DiverCorr) containing 1M image pairs to support the study of our method’s scalable generalization. 
All evaluation protocols remain the same as in the correspondence pruning task above. 

\mparagraph{DiverseCorr}
We build our DiverseCorr dataset by collecting 1M image pairs from multiple large-scale datasets~\cite{yeshwanth2023scannet, li2018megadepth, yu2023mvimgnet, zhao2023arkittrack} with distinct characteristics.
Correspondences for each image pair are generated using geometric matcher GIM$_\text{LightGlue}$~\cite{xuelun2024gim, lindenberger2023lightglue} with our GeneralPruner (${\dagger}$).
This dataset enables the analysis of scalable generalization in our pre-training paradigm (${\star}$). 

\mparagraph{Cross-Scene}
We evaluate four unseen scene splits from the outdoor benchmark~\cite{Thomee2016}, including \textit{Buckingham Palace} (BUC), \textit{Notre Dame Front} (NOT), \textit{Reichstag} (REI), and \textit{Sacré Coeur} (SAC). 
All these unseen scenes belong to the same domain (outdoor landmarks) as the training data but vary in difficulty. 
Specifically, the proportions of image pairs with an outlier ratio above 90\% are $67.8$\%, $56.3$\%, $41.7$\%, and $61.3$\% for BUC, NOT, REI, and SAC, respectively. 
As shown in Table~\ref{tab:crossscene}, our CorrFormer notably improves performance on the most challenging scenes, \ie, BUC and SAC, achieving AUC@5\textdegree~gains of $15.19$\% and $15.23$\%, respectively. 
Furthermore, our pre-training method consistently enhances performance across all scenes, including both challenging and relatively easy ones (\eg, $6.47$\%, $9.20$\%, and $5.27$\% improvements on BUC, NOT, and REI). 
When additional data are incorporated into pre-training, our approach attains further improvements across all scenes. 
These cross-scene evaluations reveal two key insights, 
\textbf{1)} Our encoder demonstrates superior generalization capability, achieving more robust performance on challenging unseen scenes compared with prior methods. 
\textbf{2)} Our geometry-consistent pre-training method (${\dagger}$) is highly versatile, improving generalization across both easy and difficult scenes. 
With additional data (${\star}$), it further exhibits stronger scalability in generalization. 

\begin{table}[t]
\caption{Generalization comparison results on four unseen outdoor scenes~\cite{Thomee2016} for camera pose estimation. }
\setlength{\tabcolsep}{3pt}
\renewcommand\arraystretch{1.}
\centering
\vspace{-0.2cm}
\begin{threeparttable}
\scalebox{0.82}{
\begin{tabular}{l|c|c|c|c}
\toprule
\multirow{2}{*}{\textbf{Method}}  & \multicolumn{4}{c}{\textbf{Pose Estimation (AUC@5\textdegree~/ @20\textdegree)}}
\\\cmidrule{2-5}
& \textbf{BUC} & \textbf{NOT} & \textbf{REI} & \textbf{SAC}
\\\midrule\midrule
CLNet~\cite{Zhao2021}  & $16.46$ / $59.87$ & $17.47$ / $56.43$ & $32.93$ / $74.64$ & $33.66$ / $70.66$ 
\\
PGFNet~\cite{liu2023pgfnet} & $13.52$ / $56.63$ & $19.44$ / $58.83$ & $36.49$ / $77.99$ & $27.17$ / $66.94$ 
\\
ConvMatch~\cite{convmatch2023} & $16.02$ / $61.64$ & $23.03$ / $61.96$ & $37.36$ / $78.34$ & $30.16$ / $69.58$
\\
NCMNet~\cite{liu2023ncm}  & $23.77$ / $68.32$ & $25.35$ / $64.04$ & $45.13$ / $80.81$ & $44.30$ / $76.78$ 
\\
MGNet~\cite{dai2024mgnet}  & $20.29$ / $65.47$ & $21.54$ / $62.16$ & $43.63$ / $81.03$ & $43.48$ / $78.09$ 
\\
BCLNet~\cite{miao2024bclnet}  & $25.76$ / $69.24$ & $26.32$ / $64.41$ & $45.35$ / $81.33$ & $45.23$ / $77.73$ 
\\
DeMatch~\cite{zhang2024dematch} & $17.04$ / $62.28$ & $25.81$ / $64.99$ & $42.18$ / $80.94$ & $38.14$ / $73.41$
\\\midrule\midrule
CorrFormer & $27.38$ / $70.89$ & $27.07$ / $65.79$ & $44.93$ / $80.40$ & $52.12$ / \underline{$82.09$}
\\
\myhookarrow w. CorrMAE$^{\dagger}$ & \underline{$29.15$} / \underline{$71.59$} & \underline{$29.56$} / \underline{$68.10$} & \underline{$47.30$} / \underline{$81.45$} & \underline{$52.45$} / $81.94$
\\
\myhookarrow w. CorrMAE$^{\star}$ & $\mathbf{32.48}$ / $\mathbf{74.13}$ & $\mathbf{30.64}$ / $\mathbf{70.30}$ & $\mathbf{48.27}$ / $\mathbf{82.36}$ & $\mathbf{52.94}$ / $\mathbf{82.42}$
\\
\bottomrule
\end{tabular}}
\end{threeparttable}
\label{tab:crossscene}
\vspace{-0.3cm}
\end{table}

\begin{table*}[t]
\caption{Generalization comparison results for camera pose estimation on the zero-shot cross-domain benchmark~\cite{xuelun2024gim}.}
\setlength{\tabcolsep}{3pt}
\renewcommand\arraystretch{1.}
\centering
\vspace{-0.2cm}
\begin{threeparttable}
\scalebox{1}{
\begin{tabular}{l|c|cccccccc|cccc}
\toprule
\multirow{2}{*}{\textbf{Method}} & \multirow{2}{*}{\textbf{Mean}} & \multicolumn{8}{c|}{\textbf{Real}} & \multicolumn{4}{c}{\textbf{Simulate}} 
\\\cmidrule{3-14}
& & \textbf{GL3}  & \textbf{BLE}  & \textbf{ETI}  & \textbf{ETO}  & \textbf{KIT} & \textbf{WEA} & \textbf{SEA} & \textbf{NIG} & \textbf{MUL} & \textbf{SCE} & \textbf{ICL} & \textbf{GTA}
\\\midrule\midrule
NoPruner~\cite{Lowe2004} & $5.94$ & $5.02$ & $3.30$ & $4.68$ & $4.26$ & $15.94$ & $7.13$ & $8.95$ & $4.98$ & $9.74$ & $0.82$ & $0.55$ & $5.94$
\\  
CLNet~\cite{Zhao2021}    & $13.94$   & $15.52$   & $13.76$    & $11.14$   & $16.75$    & $21.04$   & $17.90$  & $27.69$ & $12.08$   & $10.73$  & $3.27$   & $5.77$  & $11.61$ 
\\  
PGFNet~\cite{liu2023pgfnet} & $14.25$ & $12.80$   & $11.30$    & $17.37$   & $20.70$    & $22.65$   & $16.07$  & $24.96$ & $11.33$   & $14.13$  & $2.98$   & $4.64$  & $12.10$ 
\\
ConvMatch~\cite{convmatch2023}
& $13.99$ & $11.83$   & $11.57$    & $10.35$   & $12.36$    & $22.80$   & $17.12$  & $28.23$ & $11.99$   & $\mathbf{20.64}$  & $4.88$   & $6.21$  & $9.90$  
\\
NCMNet~\cite{liu2023ncm} & $13.69$ & $14.17$   & $14.43$    & $10.38$   & $13.23$    & $23.92$   & $19.08$  &  $29.74$ & $12.82$   & $8.11$  & $4.14$   & $7.51$  & $6.72$   
\\
MGNet~\cite{dai2024mgnet} & $13.50$  & $14.40$   & $13.58$    & $10.25$   & $12.14$    & $21.21$   & $17.27$  & $26.41$ & $12.05$  & $13.51$  & $3.70$  & $6.48$  & $10.99$  
\\
BCLNet~\cite{miao2024bclnet} & $14.42$ & $17.00$   & $15.23$    & $12.84$   & $15.51$    & $23.26$   & $18.66$  & $28.17$ & $11.94$   & $7.83$  & $4.64$   & $7.48$  & $10.03$  
\\
DeMatch~\cite{zhang2024dematch} & $11.74$ & $10.00$ & $9.77$   & $7.99$    & $8.63$   & $20.17$    & $15.69$   & $22.38$  & $11.21$ & $15.00$   & $4.38$  & $5.80$   & $9.81$ 
\\\midrule\midrule
CorrFormer  & $15.14$   & $19.30$    & \underline{$18.76$}   & $13.28$    & $15.63$   & \underline{$24.76$}  & $15.94$ & $25.47$   & $11.89$  & $14.32$   & $\mathbf{6.13}$  & $\mathbf{9.94}$  & $6.26$
\\
\myhookarrow w. CorrMAE$^{\dagger}$  & $16.35$   & $19.00$    & $17.34$   & $16.54$    &  $22.52$   & $24.57$  & \underline{$19.23$} & \underline{$28.73$}   & $\mathbf{13.76}$  & $7.41$   & $5.33$  & $8.96$  & $12.78$
\\
\myhookarrow w. CorrMAE$^{\star}_1$  & \underline{$16.88$}   & \underline{$20.00$}    & $17.82$  &  \underline{$18.68$}   &  \underline{$23.15$}  & $23.71$  & $18.93$  & $28.42$   & \underline{$13.25$}  & $10.38$  & $4.67$  & \underline{$9.31$} &  \underline{$14.25$}
\\
\myhookarrow w. CorrMAE$^{\star}$  & $\mathbf{19.69}$   & $\mathbf{24.35}$    & $\mathbf{18.97}$   & $\mathbf{23.35}$    & $\mathbf{28.32}$   & $\mathbf{25.82}$  & $\mathbf{20.60}$ & $\mathbf{32.60}$   & $12.70$  & \underline{$18.09$}   & \underline{$5.23$}  & $8.91$  & $\mathbf{17.33}$
\\
\bottomrule
\end{tabular}}
\end{threeparttable}
\label{tab:crossdomain}
\end{table*}

\mparagraph{Cross-Domain}
To assess cross-domain generalization, we adopt the zero-shot evaluation benchmark~\cite{xuelun2024gim}, which covers $12$ datasets spanning diverse domains, such as driving scenes, aerial imagery, and varying weather conditions. 
We also confirm that our DiverseCorr has no overlap with this benchmark. 
As shown in Table~\ref{tab:crossdomain}, our CorrFormer outperforms recent methods in mean AUC@5\textdegree, benefiting from its stronger architectural design. 
Furthermore, our pre-training method (${\dagger}$) further improves the mean AUC@5\textdegree~score with a relative gain of $7.99$\%. 
When additional data are introduced during pre-training (${\star}$), our model achieves a substantial boost in cross-domain generalization, yielding an additional $20.43$\% improvement. 
Removing the distribution loss (the subscript-$1$ variant) during pre-training noticeably degrades cross-domain performance, confirming its importance for generalization. 
These results highlight two main insights, 
\textbf{1)} Our encoder exhibits stronger generalization across diverse domains. 
\textbf{2)} Our pre-training approach further enhances cross-domain generalization, especially when supported by additional data. 
Notably, our method achieves particularly large improvements on the GL3, ETI, ETO, and GTA datasets, owing to the diverse scene patterns captured in DiverseCorr. 
It suggests that our geometry-consistent pre-training paradigm provides a scalable and effective solution for improving model generalization across domains, particularly when trained on abundant, readily available data. 
However, our performance on the Simulate dataset is relatively less consistent, mainly because no simulated data are involved during the entire training process. 

\begin{figure}[t]
\vspace{-0.2cm}
\begin{center}
    \includegraphics[width=0.9\linewidth]{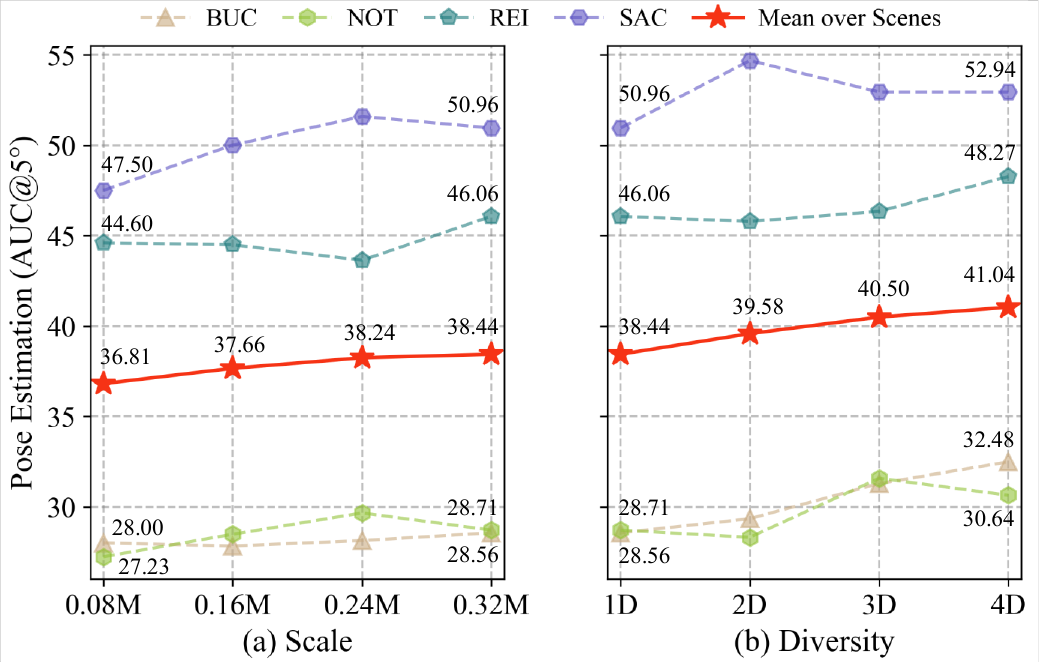}
    \vspace{-0.2cm}
    \caption{Generalization across pre-training data scale and diversity. 
    Mean performance over scenes is highlighted.}\label{fig:scalability}
\end{center}
\vspace{-0.4cm}
\end{figure}

\mparagraph{Scale and Diversity}
We study how pre-training data scale and diversity affect generalization. 
In all experiments, we pre-train on subsets of DiverseCorr, fine-tune on correspondence pruning, and evaluate cross-scene generalization. 
\textbf{1) Dataset scale.} To isolate scale from diversity, we construct four indoor-only subsets (0.08M-0.32M). 
As shown in Figure~\ref{fig:scalability}(a), performance generally improves as more pre-training data are used. 
With small subsets, the gains are limited and can be worse than fully supervised CorrFormer, suggesting that low-scale pre-training is more vulnerable to outliers. 
Once the dataset reaches a sufficient size (around 0.2M), CorrMAE starts to deliver clear and consistent improvements. 
\textbf{2) Dataset diversity.} We further build four subsets with increasing diversity by progressively adding different scene types (indoor, outdoor, dynamic, and object-centric). 
As shown in Figure.~\ref{fig:scalability}(b), diversity is important for generalization. 
Adding diverse scenes (1D-3D) brings steady gains, while adding the more homogeneous object-centric split (4D) yields smaller improvements. 

\subsection{Further Analysis}
\label{sec:analysis}
In this section, we perform in-depth analyses on compatibility, efficiency, and threshold sensitivity to further demonstrate the practical advantages of our method. 

\begin{table}[t]
\caption{Quantitative comparison of the recent matcher~\cite{sun2021loftr} paired with different pruners on the outdoor benchmark~\cite{Thomee2016}.}
\setlength{\tabcolsep}{3pt}
\renewcommand\arraystretch{1.}
\centering
\vspace{-0.2cm}
\begin{threeparttable}
\scalebox{0.9}{
\begin{tabular}{l|p{85pt}<{\centering}|p{89pt}<{\centering}}
\toprule
\multirow{2}{*}{\textbf{Method}}  & \multicolumn{2}{c}{\textbf{Pose Estimation (AUC@5\textdegree~/ @10\textdegree~/ @20\textdegree)}}
\\\cmidrule{2-3}
& \textbf{DLT~\cite{hartley1997defense}} & \textbf{RANSAC~\cite{Fischler1981}}
\\\midrule\midrule
XFeat+NN~\cite{potje2024xfeat}     & / & $13.71$ / $28.65$ / $46.85$
\\
\myhookarrow w. OANet~\cite{Zhang2019}     & $10.18$ / $24.20$ / $43.40$ & $23.81$ / $43.24$ / $62.09$ 
\\
\myhookarrow w. CLNet~\cite{Zhao2021}       & $14.14$ / $31.18$ / $52.06$ & $25.25$ / $45.51$ / $64.68$
\\
\myhookarrow w. NCMNet~\cite{liu2023ncm}    & $18.52$ / $38.53$ / $59.73$ & $27.60$ / $49.02$ / $67.92$ 
\\
\myhookarrow w. GeneralPruner$^{\star}$  & $\mathbf{20.48}$ / $\mathbf{40.75}$ / $\mathbf{61.45}$ & $\mathbf{29.24}$ / $\mathbf{50.41}$ / $\mathbf{68.64}$
\\\midrule\midrule
LoFTR~\cite{sun2021loftr}    & / & $40.89$ / $61.29$ / $76.80$ 
\\
\myhookarrow w. OANet     & $24.43$ / $40.89$ / $59.12$ & $37.56$ / $57.92$ / $74.14$ 
\\
\myhookarrow w. CLNet      & $34.18$ / $52.71$ / $69.49$ & $42.19$ / $62.09$ / $77.11$ 
\\
\myhookarrow w. NCMNet    & $35.42$ / $53.86$ / $70.47$ & $43.52$ / $63.08$ / $77.83$ 
\\
\myhookarrow w. GeneralPruner$^{\star}$  & $\mathbf{41.32}$ / $\mathbf{60.83}$ / $\mathbf{76.09}$ & $\mathbf{46.10}$ / $\mathbf{65.62}$ / $\mathbf{79.77}$ 
\\
\bottomrule
\end{tabular}}
\end{threeparttable}
\caption*{\scriptsize Putative correspondences generated by LoFTR and XFeat are used as inputs.}
\label{tab:loftr}
\vspace{-0.3cm}
\end{table}

\begin{figure}[t]
\vspace{-0.2cm}
\begin{center}
    \includegraphics[width=0.9\linewidth]{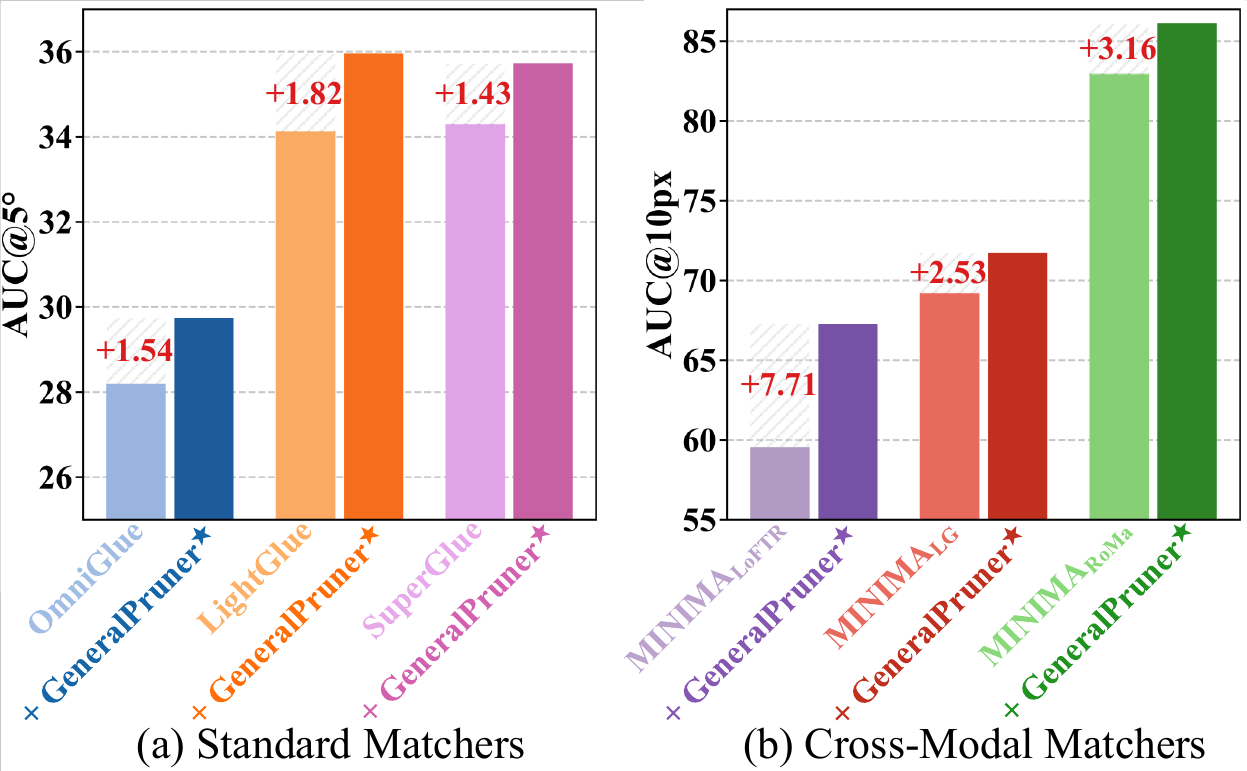}
    \vspace{-0.2cm}
    \caption{Plug-and-play evaluation on (a) standard~\cite{jiang2024omniglue, lindenberger2023lightglue, Sarlin2020} and (b) cross-modal~\cite{ren2025minima} matchers. 
    Evaluations are conducted on the outdoor benchmark~\cite{Thomee2016} and the Map-Visible dataset~\cite{wu2025mapglue}.}\label{fig:diff_matcher}
\end{center}
\vspace{-0.4cm}
\end{figure}

\begin{figure}[t]
\begin{center}
    \includegraphics[width=\linewidth]{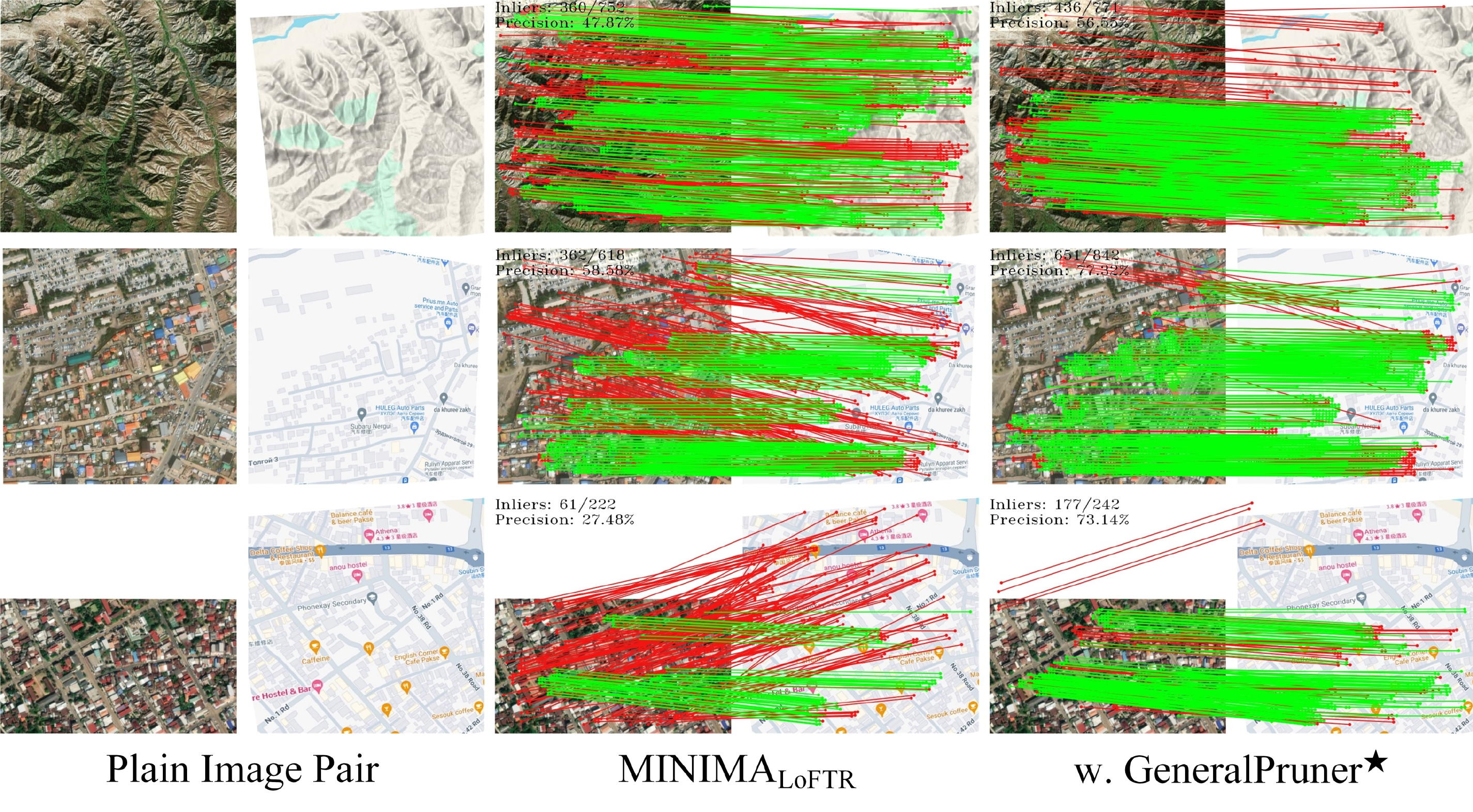}
    \caption{Partial typical visualization results on the Visible-Map dataset~\cite{wu2025mapglue}.}\label{fig:cross_modal_vis}
\end{center}
\vspace{-0.3cm}
\end{figure}

\mparagraph{Compatibility}
In this part, we analyze the compatibility of our method from three practical aspects, \ie, retraining-based, standard plug-and-play, and cross-modal plug-and-play matchers. 
\textbf{1) Retraining-based Compatibility.} 
We follow the retraining and evaluation protocols of previous works~\cite{liu2023ncm, zhang2024dematch}. 
Specifically, putative correspondences are generated by the recent matcher LoFTR~\cite{sun2021loftr} and recent SoTA feature detector XFeat~\cite{potje2024xfeat}. 
We then retrain each pruner to replace LoFTR’s built-in filtering strategy, with the epipolar distance threshold adjusted to $1\times10^{-7}$. 
After retraining, all pruners are evaluated on the outdoor benchmark~\cite{Thomee2016} under both DLT and RANSAC estimators. 
As shown in Table~\ref{tab:loftr}, nearly all pruners outperform the vanilla LoFTR and XFeat, and our GeneralPruner even achieves superior performance with the efficient DLT, 
surpassing LoFTR ($41.32$ vs. $40.89$ AUC@5\textdegree) and XFeat ($20.48$ vs. $13.71$ AUC@5\textdegree) with RANSAC. 
When combined with RANSAC, our approach further improves LoFTR and XFeat by $5.21$ and $15.53$ in AUC@5\textdegree. 
These results indicate that when a matcher is expected to generalize to unseen scenes or datasets, 
training a lightweight pruner serves as an efficient alternative to retraining a heavy matcher, while allowing fast adaptation with minimal computational cost. 
\textbf{2) Plug-and-Play Compatibility with Standard Matchers.} 
Following the setups of NCMNet~\cite{liu2023ncm} and DeMatch~\cite{zhang2024dematch}, we integrate our GeneralPruner with recent SoTA matchers, 
including OmniGlue~\cite{jiang2024omniglue}, LightGlue~\cite{lindenberger2023lightglue}, and SuperGlue~\cite{Sarlin2020}. 
As shown in Figure~\ref{fig:diff_matcher} (a), our method consistently boosts performance on the outdoor benchmark, improving pose estimation accuracy by $5.46$\%, $5.33$\%, and $4.17$\% on these matchers, respectively. 
This demonstrates that our approach can function as a flexible plug-and-play component, seamlessly enhancing existing matching methods. 
\textbf{3) Plug-and-Play Compatibility with Cross-Modal Matchers.} 
To further assess robustness under modality shifts, we integrate GeneralPruner with recent cross-modal matchers~\cite{ren2025minima} and conduct zero-shot homography estimation on an unseen modality dataset~\cite{wu2025mapglue}, \ie, visible light image-electronic navigation map (Visible-Map). 
We follow the evaluation protocols of MINIMA~\cite{ren2025minima} for fair comparison. 
As shown in Figure~\ref{fig:diff_matcher} (b), our method improves the AUC@10px scores of MINIMA$_\text{LoFTR}$~\cite{sun2021loftr}, MINIMA$_\text{LG}$~\cite{lindenberger2023lightglue}, and MINIMA$_\text{RoMa}$~\cite{edstedt2024roma} by $12.94$\%, $3.66$\%, and $3.81$\%, respectively. 
These results confirm that our GeneralPruner effectively enhances cross-modal matchers under extremely challenging zero-shot generalization settings, providing reliable support for future cross-modal matching research. 
Additionally, qualitative visualizations of zero-shot cross-modal matching results are presented in Figure~\ref{fig:cross_modal_vis}. 

\begin{table}[t]
\caption{Computational efficiency comparison between our methods and recent SoTA pruners.}
\setlength{\tabcolsep}{3pt}
\renewcommand\arraystretch{1.}
\centering
\vspace{-0.2cm}
\begin{threeparttable}
\scalebox{0.9}{
\begin{tabular}{l|p{40pt}<{\centering}|p{40pt}<{\centering}p{40pt}<{\centering}p{40pt}<{\centering}p{40pt}<{\centering}}
\toprule
\textbf{Method} & \textbf{AUC@5\textdegree} & \textbf{Runtime} & \textbf{Memory} & \textbf{Parameter} 
\\\midrule\midrule
CLNet~\cite{Zhao2021}       & $25.28$ & $16.19$ & $97.09$ & $1.27$ 
\\
MS$^2$DGNet~\cite{Dai2022}  & $20.61$ & $12.69$ & $104.26$ & $2.61$
\\
ConvMatch~\cite{convmatch2023}  & $26.83$ & $18.17$ & $64.03$ & $7.49$ 
\\
NCMNet~\cite{liu2023ncm}    & $34.51$ & $47.63$ & $129.24$ & $4.77$ 
\\
UMatch~\cite{li2024u}      & $30.84$ & $38.16$ & $64.16$ & $7.76$ 
\\
BCLNet~\cite{miao2024bclnet}   & $35.70$ & $32.30$ & $93.81$ & $4.87$ 
\\
DeMatch~\cite{zhang2024dematch}   & $30.91$ & $25.40$ & $52.80$ & $5.86$ 
\\\midrule\midrule 
GeneralPruner$^{\star}$ (Small)  & $37.51$ & $27.05$ & $88.48$ & $2.51$
\\
\rowcolor{lightgray!20} \myhookarrow \textit{vs. BCLNet}  & \footnotesize{{$\textcolor[RGB]{125,119,34}{\mathbf{\uparrow 1.81}}$}} & \footnotesize{{$\textcolor[RGB]{125,119,34}{\mathbf{\downarrow 5.25}}$}} & \footnotesize{{$\textcolor[RGB]{125,119,34}{\mathbf{\downarrow 5.33}}$}} & \footnotesize{{$\textcolor[RGB]{125,119,34}{\mathbf{\downarrow 2.36}}$}}
\\
GeneralPruner$^{\star}$ (Base) & $41.04$ & $41.02$ & $94.84$ & $4.10$
\\
\rowcolor{lightgray!20} \myhookarrow \textit{vs. NCMNet}  & \footnotesize{{$\textcolor[RGB]{125,119,34}{\mathbf{\uparrow 6.53}}$}} & \footnotesize{{$\textcolor[RGB]{125,119,34}{\mathbf{\downarrow 6.61}}$}} & \footnotesize{{$\textcolor[RGB]{125,119,34}{\mathbf{\downarrow 34.40}}$}} & \footnotesize{{$\textcolor[RGB]{125,119,34}{\mathbf{\downarrow 0.67}}$}}
\\
\bottomrule
\end{tabular}}
\end{threeparttable}
\caption*{\scriptsize The units for runtime, memory, and parameters are milliseconds (ms), megabytes (MB), and millions (M), respectively.}
\label{tab:efficiency}
\vspace{-0.3cm}
\end{table}

\mparagraph{Efficiency}
We evaluate the computational efficiency of our method (small and base variants) and recent SoTAs on an NVIDIA RTX 3090 GPU, following the same protocol as~\cite{chen2021learning}. 
Two variants of our model are evaluated, where the number of transformer layers in the encoder is set to $2$ (Small) and $4$ (Base), respectively. 
As shown in Table~\ref{tab:efficiency}, our Small variant surpasses all recent methods, achieving higher accuracy with lower computational costs. 
Compared with BCLNet~\cite{miao2024bclnet}, it not only achieves better performance but also reduces overall computational cost, with a particularly notable reduction of $16.25$\% in runtime. 
Furthermore, our Base variant outperforms NCMNet~\cite{liu2023ncm}, which has a comparable computational cost, with a relative gain of $18.92$\% in AUC@5\textdegree. 
These results demonstrate that our method achieves a favorable balance between performance and efficiency, making it highly suitable for practical deployment. 

\begin{figure}[t]
\begin{center}
    \includegraphics[width=0.9\linewidth]{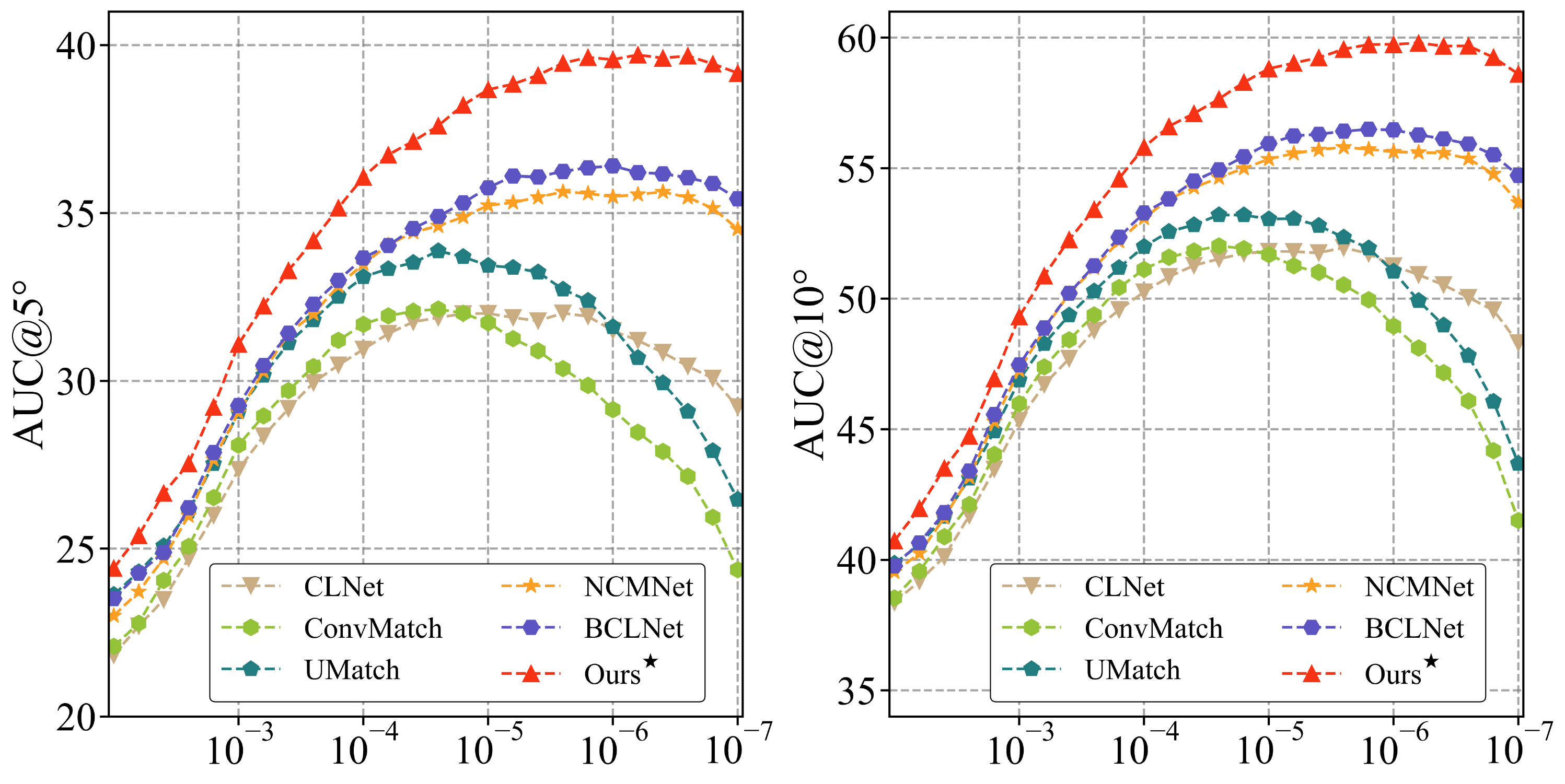}
    \vspace{-0.2cm}
    \caption{Threshold sensitivity analysis on the outdoor benchmark~\cite{Thomee2016} for pose estimation. }\label{fig:thres}
\end{center}
\vspace{-0.4cm}
\end{figure}

\mparagraph{Threshold}
In practice, the threshold is also a crucial factor affecting the performance of pruners, and analyzing their sensitivity to threshold variation is important for practical deployment. 
Following the experimental setup of UMatch~\cite{li2024u}, we evaluate recent SoTAs and ours on the outdoor benchmark~\cite{Thomee2016} under RANSAC-based post-processing to assess the sensitivity to epipolar distance thresholds. 
We report the AUC@5\textdegree~and AUC@10\textdegree~scores for the pose estimation task under different thresholds. 
As shown in Figure~\ref{fig:thres}, across a wide range of thresholds (from $1\times10^{-2}$ to $1\times10^{-7}$, covering $26$ values), our method consistently outperforms recent SoTAs. 
It achieves an optimal balance between accuracy and stability within the range of $1\times10^{-5}$ to $4\times10^{-7}$. 
These results demonstrate that our method attains a favorable trade-off between threshold sensitivity and performance, making it well-suited for practical applications. 

\section{Conclusion}   \label{sec:conclusion}
We propose a geometry-consistent pre-training paradigm for scalable and generalizable correspondence pruning. 
Based on this paradigm, CorrMAE learns robust and transferable representations via masked inlier reconstruction, avoiding outlier interference. 
We further introduce a dual-stream encoder with built-in consensus interaction, offering a simple and extensible architecture for correspondence learning. 
Extensive experiments on five downstream tasks have demonstrated the efficacy of our GeneralPruner.

\begin{figure}[t]
\begin{center}
    \includegraphics[width=\linewidth]{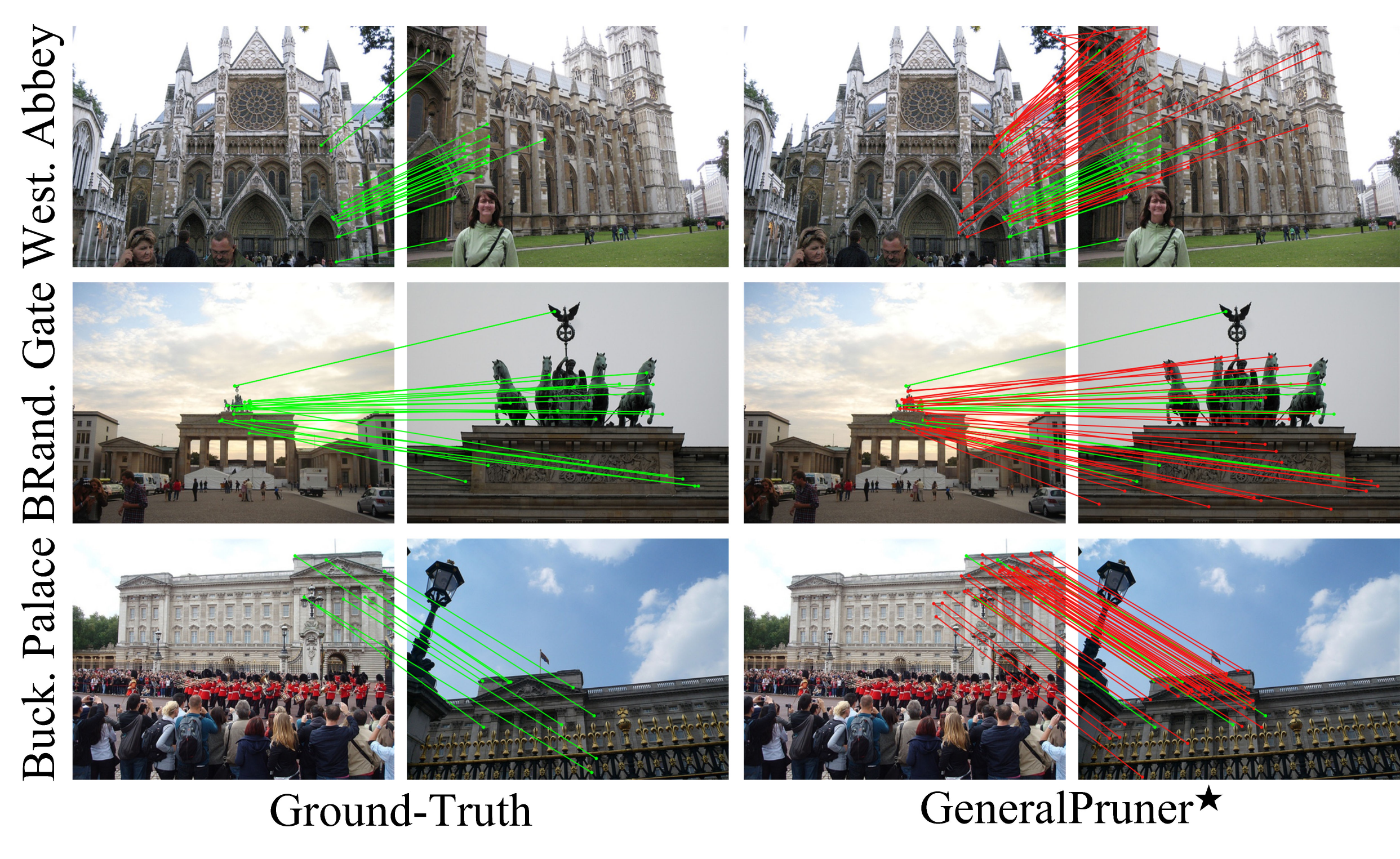}
    \caption{Failure cases under a small 3D overlap. Sparse inliers are overwhelmed by dominant outliers, making reliable pruning difficult.}\label{fig:failure}
\end{center}
\vspace{-0.4cm}
\end{figure}

\mparagraph{Limitations}
In spite of the state-of-the-art performance, there are still some limitations in our work. 
\textbf{First,} GeneralPruner is most valuable in challenging scenes with strong ambiguity, such as complex motion or structure, or cross-modal shifts. 
In clean scenes where modern end-to-end matchers already produce reliable correspondences, the additional gain from pruning can be smaller. 
In this case, using the pruner as a lightweight domain adapter is a promising direction. 
\textbf{Second,} the method depends on the putative correspondences produced by the front-end matcher. 
When dense matchers are used, the large number of correspondences can increase the pruning cost. 
Therefore, the downsampling rate should be carefully chosen to balance performance and efficiency. 
\textbf{Third,} performance may degrade in scenes with very small 3D overlap, where only a few inliers exist. 
Under such conditions, identifying the sparse inliers from dominant outliers remains challenging for the pruner, see failure cases in Figure~\ref{fig:failure}.



\ifCLASSOPTIONcaptionsoff
  \newpage
\fi



\bibliographystyle{IEEEtran}
\bibliography{refs.bib}
%

%




\end{document}